\documentstyle[epsf, 11pt]{article}

\newtheorem{remark}{Remark}

\begin{document}

\begin{titlepage}
\title{\vspace{22 mm} Camera Calibration: a USU Implementation\footnote{Not for public release. This work is supported in part by U.S. Army Automotive and Armaments Command (TACOM) Intelligent Mobility Program (Agreement no. DAAE07-95-3-0023).}}
\author{Lili Ma, YangQuan Chen, and Kevin L. Moore\\Center for Self-Organizing and Intelligent Systems\\Department of Electrical and Computer Engineering\\Utah State University\\Email: {\tt lilima@cc.usu.edu}}
\maketitle{}
\vspace{8 mm}
\begin{abstract}
The task of camera calibration is to estimate the intrinsic and extrinsic parameters of a camera model. Though there are some restricted techniques to infer the 3-D information about the scene from uncalibrated cameras, effective camera calibration procedures will open up the possibility of using a wide range of existing algorithms for 3-D reconstruction and recognition. The applications of camera calibration include vision-based metrology, robust visual platooning and visual docking of mobile robots where the depth information is important. 
\\\\
\noindent {\bf Key Words:} Camera calibration, projective model, radial distortion, radial undistortion, flexible setup, nonlinear optimization.
\end{abstract}
\pagenumbering{Roman}
\setcounter{page}{0}
\end{titlepage}

\clearpage
\tableofcontents
\clearpage

\newcommand{\indentintabular}{$\;$$\;$$\;$$\;$$\;$$\;$$\;$}

{\small
\begin{table}[htb]
\centering
\caption{List of Variables}
\label{table: variables used in this report}
\bigskip
{\begin {tabular}{|c|l|}\hline
{\bf Variable} & {\bf Explanation} \\[1ex]\hline
$n$	 					& Number of feature points in each image\\[1ex]\hline
$N$ 						& Number of images taken for calibration \\[1ex]\hline
$f$ 						& Focal length \\[1ex]\hline
$\theta$ 					& Skewness angle between two image axes \\[1ex]\hline
$P^w = [X^w, Y^w, Z^w]^T$ 		& 3-D point in world reference frame\\[1ex]\hline
$P^c = [X^c, Y^c, Z^c]^T$ 		& 3-D point in camera reference frame \\[1ex]\hline
$p=[x^c, \ y^c]^T$ 			& 2-D point in camera frame with $z = f$ \\[1ex]\hline
$M_i = [X^w, Y^w, 1]^T$ 		& 3-D point in world reference frame with $Z^w = 0$ \\[1ex]\hline
$m_i$ 					& The corresponding projected point in image plane of $M_i$ \\[1ex]\hline
$\alpha, \beta, \gamma, u_0, v_0$	& 5 intrinsic parameters \\[1ex]\hline
$(s_x, \ s_y)$				& Effective sizes of the pixel in the horizontal and vertical directions \\[1ex]\hline
$(f_x, \ f_y)$ 				& $f_x = f/s_x, \ f_y = f/s_y$\\[1ex]\hline
${\bf k} = (k_1, \ k_2)$ 		& Distortion coefficients \\[1ex]\hline
$(u_d, \ v_d)$ 				& Real observed distorted image points \\[1ex]\hline
$(u, \ v)$ 					& Ideal projected undistorted image points \\[1ex]\hline 
$(x, \ y)$ 					& $\left[ \matrix{x \cr y \cr 1}\right] = 1/f \left[ \matrix{x^c \cr y^c \cr f}\right] = A^{-1} \left[ \matrix{u \cr v \cr 1}\right]$ \\[4ex]\hline
$(x', \ y')$				& $\left[ \matrix{x'\cr y' \cr 1}\right] = A^{-1} \left[ \matrix{u_d \cr v_d \cr 1}\right]$\\[3.2ex]\hline
$\lambda$ 					& Scaling factor \\[1ex]\hline
$J$ 						& Objective function \\[1ex]\hline 
${\bf A} = \left [ \matrix {
\alpha & \gamma &u_0 \cr
0 & \beta & v_0 \cr
0 & 0 & 1
} \right ]$ 				& Camera intrinsic matrix \\[4ex]\hline
$\bf B = {\bf A}^{-T} {\bf A}^{-1}$ & Absolute conic \\[1ex]\hline
$\bf R$ 					& $3 \times 3$ rotation matrix \\[1ex]\hline
$\bf t$ 					& $3 \times 1$ translation vector \\[1ex]\hline
$[{\bf R} \ {\bf t}]$ 			& Camera extrinsic matrix \\[1ex]\hline
$\bf H$ 					& Homography matrix \\[1ex]\hline
$\bf L$ 					& Matrix used to estimate homography matrix $\bf H$ \\[1ex]\hline
$\bf V$ 					& Matrix stacking constraints to estimate intrinsic parameters \\[1ex]\hline
\end {tabular}}
\end{table}}

\clearpage
\section{Introduction}
\pagenumbering{arabic}
Depending on what kind of calibration object being used, there are mainly two categories of calibration methods: {\bf {\it photogrammetric calibration}} and {\bf {\it self-calibration}}. Photogrammetric calibration refers to those methods that observe a calibration object whose geometry in 3-D space is known with a very good precision. Self-calibration does not need a 3-D calibration object. Three images of a coplanar object taken by the same camera with fixed intrinsic camera parameters are sufficient to estimate both intrinsic and extrinsic parameters. The obvious advantage of the self-calibration method is that it is easy to set up and the disadvantage is that it is usually considered unreliable. However, the author of \cite{Richard97indefenseof8-pointalgorithm} shows that by preceding the algorithm with a very simple normalization (translation and rotation of the coordinates of the matched points), results are obtained comparable with the best iterative algorithms. A four step calibration procedure is proposed in \cite{Heikkil97fourstepcameracalibration}. The four steps are: linear parameter estimation, nonlinear optimization, correction using circle/ellipse, and image correction. But for a simple start, linear parameter estimation and nonlinear optimization are enough. In \cite{STURM99planebasedcalibrationsigularities}, a plane-based calibration method is described where the calibration is performed by first determining the absolute conic ${\bf B} = {\bf A}^{-T} {\bf A}^{-1}$, where $A$ is a matrix formed by a camera's 5 intrinsic parameters (See Section \ref{section: intrinsic parameters}). In \cite{STURM99planebasedcalibrationsigularities}, the parameter $\gamma$ (a parameter describing the skewness of the two image axes) is assumed to be zero and the authors observe that only the relative orientations of planes and camera is of importance for singularities and planes that are parallel to each other provide exactly the same information. Recently, Intel distributes its ``Camera Calibration Toolbox for Matlab'' freely available online \cite{intel}. The Intel camera calibration toolbox first finds the feature locations of the input images, which are captured by the camera to be calibrated using a checkerboard calibration object. Then, it calculates the camera's intrinsic parameters. However, when we used the images captured by our desktop camera as the input images, the detected feature locations contain great errors. We decided not to use Intel's method since its flexibility and accuracy are poor. Therefore, in this report, our work is mainly based on the self-calibration algorithm originally developed by Microsoft Research Group \cite{zhang99calibrationinpaper,zhang99calibrationinreport}, which has been commonly regarded as a great contribution to the camera calibration. The key feature of Microsoft's calibration method is that the absolute conic $\bf B$ is used to estimate the intrinsic parameters and the parameter $\gamma$ is considered. The proposed technique in \cite{zhang99calibrationinpaper,zhang99calibrationinreport} only requires the camera to observe a planar pattern at a few (at least 3, if both the intrinsic and the extrinsic parameters are to be estimated uniquely) different orientations. Either the camera or the calibration object can be moved by hand as long as they cause no singularity problem and the motion of the calibration object or camera itself need not to be known in advance. 

By ``flexibility'', we mean that the calibration object is coplanar and easy to setup while by ``robustness'', it implies that the extracted feature locations are accurate and the possible singularities due to improperly input images can be detected and avoided.

The main contributions in this report are briefly summarized as follows:
\begin{enumerate}
\item[(1)]{A complete code platform implementing Microsoft's camera calibration algorithm has been built. (Microsoft did not release the feature location pre-processor \cite{zhang99calibrationinpaper,zhang99calibrationinreport});}
\item[(2)]{A technical error in Microsoft's camera calibration equations has been corrected (Equation (\ref{eqn: intrinsic parameters from B}));}
\item[(3)]{A new method to effectively find the feature locations of the calibration object has been used in the code. More specifically, a scan line approximation algorithm is proposed to accurately determine the partitions of a given set of points;}
\item[(4)]{A numerical indicator is used to indicate the possible singularities among input images to enhance the robustness in camera calibration (under development and to be included);}
\item[(5)]{The intrinsic parameters of our desktop camera and the ODIS camera have been determined using our code. Our calibrated results have also been cross-validated using Microsoft code;}
\item[(6)]{A new radial distortion model is proposed so that the radial undistortion can be performed analytically with no numerical iteration;}
\item[(7)]{Based on the results of this work, some new application possibilities have been suggested for our mobile robots, such as ODIS.}
\end{enumerate}

The rest of the report is arranged as follows. First, some notations and preliminaries are given, such as camera pinhole model, intrinsic parameters, and extrinsic parameters. Then, the calibration method proposed in \cite{zhang99calibrationinpaper,zhang99calibrationinreport} is re-derived with a correction to a minor technical error in \cite{zhang99calibrationinpaper,zhang99calibrationinreport}. Using this method, calibration results of 3 different cameras are presented. Finally, several issues are proposed for future investigations and possible applications of camera calibration are discussed.

\section{Camera Projection Model}
\begin{figure}[htb]
\centerline{\epsfxsize=4in \epsffile{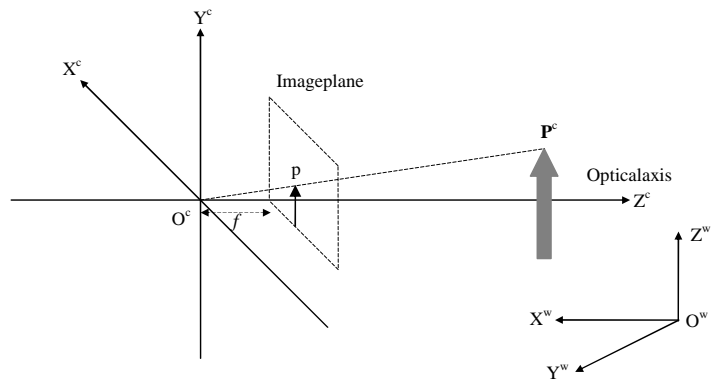}}
\caption {The perspective camera model}
\label{fig: perspective camera model}
\end{figure}

To use the information provided by a computer vision system, it is necessary to understand the geometric aspects of the imaging process, where the projection from 3-D world reference frame to image plane (2-D) causes direct depth information to be lost so that each point on the image plane corresponds to a ray in the 3-D space \cite{Seth96visualservoing}. The most common geometric model of an intensity imaging camera is the {\bf {\it perspective}} or {\bf {\it pinhole}} model (Figure \ref{fig: perspective camera model}). The model consists of the image plane and a 3-D point $O_c$, called the {\it center} or {\it focus of projection}. The distance between the image plane and $O_c$ is called the {\bf {\it focal length}} and the line through $O_c$ and perpendicular to the image plane is the {\bf {\it optical axis}}. The intersection between the image plane and the optical axis is called the {\bf {\it principle point}} or the {\bf {\it image center}}. As shown in Figure \ref{fig: perspective camera model}, the image of $P^c$ is the point at which the straight line through $O_c$ and $P^c$ intersects the image plane. The basic perspective projection \cite{Emanuele98introductorycomputervision} in the camera frame is 
\begin{eqnarray}
\label{eqn: xc, yc, Xc, Yc}
\left [ \matrix {x^c \cr y^c}\right] = \frac{f}{Z^c} \left [ \matrix {X^c \cr Y^c}\right], 
\end{eqnarray}
where $P^c = [X^c, Y^c, Z^c]^T$ is a 3-D point in the camera frame and $p = [x^c, y^c]^T$ is its projection in the camera frame. In the camera frame, the third component of an image point is always equal to the focal length $f$. For this reason, we can write $p = [x^c, y^c]^T$ instead of $p = [x^c, y^c, f]^T$. 

\section{Aspects that Real Cameras Deviate from Pinhole Model}
\begin{figure}[htb]
\centerline{\epsfxsize=2.2in \epsffile{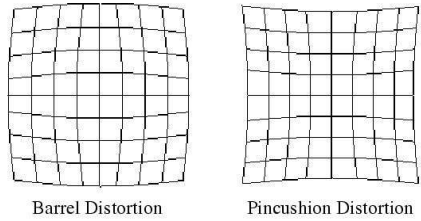}}
\caption {The barrel distortion and the pincushion distortion}
\label{fig: barrel distortion and pincushion distortion}
\end{figure}

A real camera deviates from the pinhole model in several aspects. The most significant effect is lens distortion. Because various constraints in the lens manufacturing process, straight lines in the world imaged through real lenses generally become somewhat curved in the image plane. However, this distortion is almost always radially symmetric and is referred to as the {\bf {\it radial distortion}}. The radial distortion that causes the image to bulge toward the center is called the {\bf {\it barrel distortion}}, and distortion that causes the image to shrink toward the center is called the {\bf {\it pincushion distortion}} (See Figure \ref{fig: barrel distortion and pincushion distortion}). The center of the distortions is usually consistent with the image center. 

The second deviation is the flatness of the imaging media. However, digital cameras, which have precisely flat and rectilinear imaging arrays, are not generally susceptible to this kind of distortion. 

\begin{figure}[htb]
\centerline{\epsfxsize=3.5in \epsffile{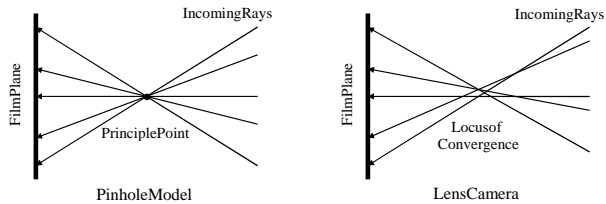}}
\caption {Lens camera deviates from the pinhole model in locus of convergence}
\label{fig: lens camera deviates from pinhole model in locus of convergence}
\end{figure}

Another deviation is that the imaged rays do not necessarily intersect at a point, which means there is not a mathematically precise principle point as illustrated in Figure \ref{fig: lens camera deviates from pinhole model in locus of convergence}. This effect is most noticeable in extreme wide-angle lenses. But the locus of convergence is almost small enough to be treated as a point especially when the objects being imaged are large with respect to the locus of convergence. 

\section{Camera Parameters}

\hrule \vskip 0.08in
\centerline {\bf Definition: Camera Parameters}
{\bf {\it Camera parameters}} are the parameters linking the coordinates of points in 3-D space with the coordinates of their corresponding image points. In particular, the {\bf {\it extrinsic parameters}} are the parameters that define the location and orientation of the camera reference frame with respect to the world reference frame and the {\bf {\it intrinsic parameters}} are the parameters necessary to link the pixel coordinates of an image point with the corresponding coordinates in the camera reference frame. 
\vskip 0.08in \hrule \vskip 0.08in 

\subsection{Extrinsic Parameters}
The extrinsic parameters are defined as any set of geometric parameters that uniquely define the transformation between the world reference frame and the camera frame. A typical choice for describing the transformation is to use a $3\times 1$ vector ${\bf t}$ and a $3 \times 3$ orthogonal rotation matrix ${\bf R}$ such that $P^c = {\bf R}P^w + {\bf t}$. According to Euler's rotation theorem, an arbitrary rotation can be described by only three parameters. As a result, the rotation matrix $\bf R$ has 3 degree-of-freedom and the extrinsic parameters totally have 6 degree of freedom. Given a rotation matrix $\bf R$ in Equation (\ref{eqn: rotation matrix}), one method to get the 3 parameters that uniquely describe this matrix is to extract $ZYZ$ Euler angles \cite{Richard94robotics}, denoted by $(a, b, c)$, such that

\begin {equation}
{\bf R} = {\left [\matrix {
r_{11} & r_{12} & r_{13}	\cr
r_{21} & r_{22} & r_{23}	\cr
r_{31} & r_{32} & r_{33}	
}\right]}
\label {eqn: rotation matrix}
\end {equation}

\begin {equation}
{\bf R} = {\bf R}_z(a) \ {\bf R}_y(b) \ {\bf R}_z(c),
\end {equation}
where
\begin {equation}
{\bf R}_z(c) = {\left [\matrix {
\cos(c) & -\sin(c) & 0	\cr
\sin(c) & \cos(c)  & 0	\cr
0 & 0 & 1	
}\right]} , \ \ \ 
{\bf R}_y(b) = {\left [\matrix {
\cos(b)& 0 & \sin(b)	\cr
0 & 1 & 0	\cr 
-\sin(b) & 0 & \cos(b)
}\right]}.
\end {equation}
When $sin(b) \neq 0$, the solutions for $(a, b, c)$ are
\begin{eqnarray}
b&=&\arctan2 \ (\sqrt{r_{31}^2 + r_{32}^2}, r_{33}), \nonumber \\
a&=&\arctan2 \ (r_{23}/\sin(b), r_{13}/\sin(b)), \\
c&=&\arctan2 \ (r_{32}/\sin(b), -r_{31}/\sin(b)). \nonumber  
\end{eqnarray}

\subsection{Intrinsic Parameters}
\label{section: intrinsic parameters}
The intrinsic parameters are as follows:

\begin{itemize}
\item {The focal length: $f$}
\item {The parameters defining the transformation between the camera frame and the image plane}\\
Neglecting any geometric distortion and with the assumption that the CCD array is made of rectangular grid of photosensitive elements, we have:
\begin{eqnarray}
\label{eqn: xc, yc, u, v}
\begin{array}{l}
x^c=-(u-u_0) \ s_x \\
y^c=-(v-v_0) \ s_y 
\end{array}
\end{eqnarray}
with $(u_0, v_0)$ the coordinates in pixel of the image center and ${s_x, s_y}$ the effective sizes of the pixel in the horizontal and vertical direction respectively. Let $f_x = f/s_x, f_y = f/s_y$, the current set of intrinsic parameters are $u_0, v_0, f_x$, and $f_y$.

\item{The parameter describing the skewness of the two image axes: $\gamma = f_y\tan{\theta}$}\\
The skewness of two image axes is illustrated in Figure \ref{fig: skewness}.
\begin{figure}[htb]
\centerline{\epsfxsize = 2.8in \epsffile{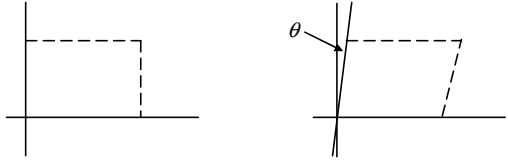}}
\caption {Skewness of two image axes}
\label{fig: skewness}
\end{figure}

\item{The parameters characterizing the radial distortion: $k_1$ and $k_2$}\\
The radial distortion is governed by the Equation \cite{Juyang92distortionmodel}
\begin{equation}
F(r) = r \ f(r) = r \ (1 + k_1 r^2 + k_2 r^4 + k_3 r^6+ \cdots).
\end{equation}
Two coefficients for distortion are usually enough. The relationship between the distorted and the undistorted image points can be approximated using
\begin{eqnarray}
\label{eqn: radial distortion}
\begin{array}{l}
u_d = u + (u-u_0) \ [k_1(x^2+y^2) + k_2(x^2+y^2)^2] \\
v_d = v + (v-v_0) \ [k_1(x^2+y^2) + k_2(x^2+y^2)^2]
\end{array}
\end{eqnarray}
where $(u_d, v_d)$ are the real observed distorted image points and $(u, v)$ the ideal projected undistorted image points. So, till now, the set of intrinsic parameters are $u_0, v_0, f_x, f_y, \gamma, k_1$, and $k_2$.
\end{itemize}

\subsection{Projection Matrix}
With homogeneous transform and the camera parameters, we can have a $3 \times 4$ matrix $\bf M$, called the {\it projection matrix}, that directly links a point in the 3-D world reference frame to its projection in the image plane. That is:
\begin{equation}
\label{eqn: projection matrix}
\lambda \left [\matrix{u \cr v \cr 1} \right ] = {\bf M} \left [\matrix{X^w \cr Y^w \cr Z^w \cr 1} \right ] = {\bf A} \left [\matrix{{\bf R} & {\bf t}}\right] \left [\matrix{X^w \cr Y^w \cr Z^w \cr 1} \right ] = \left [ \matrix { \alpha & \gamma &u_0 \cr 0 & \beta & v_0 \cr 0 & 0 & 1 } \right ] \left [\matrix{{\bf R} & {\bf t}}\right] \left [\matrix{ X^w \cr Y^w \cr Z^w \cr 1} \right ].
\end{equation}
where $\lambda$ is an arbitrary scaling factor and the matrix $\bf A$ fully depends on the intrinsic parameters. The calibration method used in this work is to first estimate the projection matrix and then use the absolute conic to estimate the intrinsic parameters \cite{zhang99calibrationinpaper,zhang99calibrationinreport}. From Equation (\ref{eqn: xc, yc, Xc, Yc}) and (\ref{eqn: xc, yc, u, v}), we have
\begin{eqnarray}
u = - \frac{f}{s_x} \frac{X^c}{Z^c} + u_0. 
\end{eqnarray}
From Equation (\ref{eqn: projection matrix}), we have 
\begin{eqnarray}
u = \alpha \frac{X^c}{Z^c} + u_0
\end{eqnarray}
with scaling factor $\lambda = Z^c$. From the above two equations, we get $\alpha = -f/s_x = -f_x$. In a same manner, $\beta = -f_y$.

\section{Extraction of Feature Locations}
\label{section: extraction of feature locations}

\subsection{Calibration Object}
\begin{figure}[htb]
\centerline{\epsfxsize=2.5in \epsffile{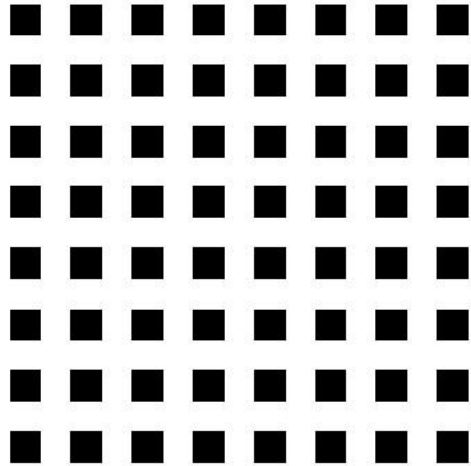}}
\caption {Current calibration object}
\label{fig: calibration object}
\end{figure}

The calibration method illustrated here is a self-calibration method, which uses a planar calibration object shown in Figure \ref{fig: calibration object}, where 64 squares are separated evenly and the side of each square is 1.3 cm. 

The procedures to extract the feature locations of the above calibration object are illustrated in Table \ref{table: procedures to extract feature locations}. The input image is an intensity image. After thresholding it with a certain value (which is 150 in our case), we can get a binary image. The binary image then goes through some Connected Component Labeling algorithm \cite{Rudyreport} \cite{Lili01connectedcomponentlabeling} that outputs a {\it region map} where each class of the connected pixels is given a unique label. For every class in the region map, we need to know whether or not it can be a square box. In our approach, this is done by first detecting the edge of each class and then finding the number of partitions of the edge points. If the number of partitions is not equal to 4, which means it is not a 4-sided polygon, we will bypass this class. Otherwise, we will fit a line using all the edge points that lie between each two adjacent partition points and thus get 4 line fits. The final output of this class is the intersections of these 4 lines that approximate the 4 corners of each box. After running through all the classes in the region map, if the number of detected boxes equals to the actual number of boxes in calibration object, we will record all the detected corners and arrange them in the same order as for the 3-D points in space (for a given calibration object, assume $Z^w = 0$, we know the exact coordinates of the feature points in world reference frame and we need to arrange these feature points in certain order so that after detecting feature points in the observed images, we can have an algorithm to seek the map from a point in the world frame to its corresponding projection in the image plane). After detecting several this kind of images, we are fully prepared to do calibration calculation. 
\begin{table}[here]
\centering
\caption{Procedures to Extract Feature Locations for One Input Image}
\label{table: procedures to extract feature locations}
\bigskip
{\begin {tabular}{||l||}\hline
Threshold input intensity image (PGM) to make it binary (the threshold is 150)\\
Find connected components using 8-connectivity method\\ \\
Loop for every class in the region map \\
\indentintabular Select the class whose area is $<$ 3000 and $>$ 20\\
\indentintabular Binary edge detection of this class\\
\indentintabular Find partitions of the edge points\\
\indentintabular If $\#$ of partitions = 4\\
\indentintabular \indentintabular Line fit between each two adjacent partition points\\
\indentintabular \indentintabular Output 4 line intersections\\
\indentintabular End if\\
End loop\\ \\
If the total $\#$ of intersections = 4 $\times$ $number\_of\_boxes\_in\_calibration\_object$\\
\indentintabular Arrange intersections in the same order as points in the world reference frame\\
End if
\\\hline
\end {tabular}}
\end{table}

\subsection{Binary Image Edge Detection (Boundary Finding)}
\begin{figure}[htb]
\centerline{\epsfxsize=2.2in \epsffile{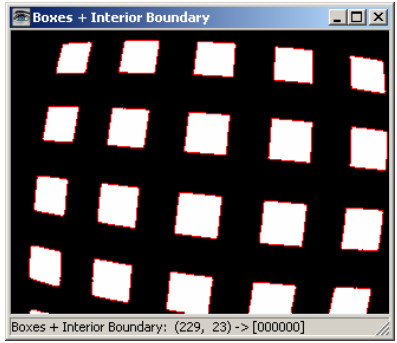}}
\caption {Objects and their interior 8-boundary}
\label{fig: 8-boundary}
\end{figure}

A boundary point of an object in a binary image is a point whose 4-neighborhood or 8-neighborhood intersects the object and its complement. Boundaries for binary images are classified by their connectivity and by whether they lie within the object or its complement. The four classifications are: interior or exterior 8-boundaries and interior or exterior 4-boundaries \cite{Gerhard00HandbookofComputerVisionAlgorithmsinImageAlgebra}. In our approach, we use interior 8-boundary operator, shown in Figure \ref{fig: 8-boundary},  which is denoted as:
\begin{equation}
{\bf b} = (1-({\bf a} \diamond N)) \ {\bf a}, 
\end{equation}
where
\begin{enumerate}
\item[(1)] {$\bf{a}$ is the input binary image}
\item[(2)] {$\bf{b}$ is the output boundary binary image}
\item[(3)] {N is the 4-neighborhood: $N(p(u,v))=\{y: y =(u \pm j, v) \ or \ y = (u, v \pm i), i, j \in \{0,1\}\}$}
\item[(4)] {For each pixel $p(u,v)$ in $\bf a$, $p(u,v) \ \diamond$ N = minimum pixel value around $p(u,v)$ in the sense of 4-neighborhood}
\end{enumerate}

\subsection{Partitions of Edge Points}
\begin{figure}[htb]
\centerline{\epsfxsize=3in \epsffile{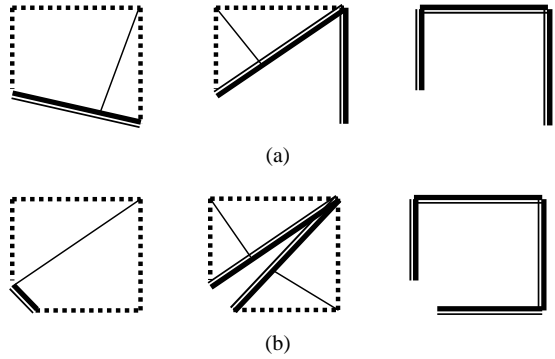}}
\caption {Illustration of scan line approximation algorithm (a) 3 sides (b) 4 sides}
\label{fig: scanlineapproximation}
\end{figure}

Given a set of points that characterize the boundary of some object, a common question is what shape this object is, when we try to use polygons, especially the convex polygons, to denote objects in the real world. The set of points can be the output of some range finding sensors, such as laser and sonar. Or, it can come from images captured by a camera and is preprocessed by some edge detector, which is just the case we are discussing. In our problem, we know beforehand that the region of interest is a square and we can use the scan line approximation method \cite{Katsoulas01scanlineapproximation,Lili02Fit} to find the number of partitions. The scan line approximation algorithm is described in Table \ref{table: scan line approximation algorithm}. Figure \ref{fig: scanlineapproximation} is an illustration.
\begin{table}[here]
\centering
\caption{Scan Line Approximation Algorithm \cite{Katsoulas01scanlineapproximation,Lili02Fit}}
\label{table: scan line approximation algorithm}
\bigskip
{\begin {tabular}{||l||}\hline
\centerline{\bf{Problem Definition}}\\
{\bf Assumption}: Object is described using a convex polygon\\
{\bf Given}: A set of data points that have already been sorted in certain order\\
{\bf Find}: Partition points\\
\centerline{\bf{Algorithm}}\\
{\bf Scan$\_$Line$\_$Approximation} (start$\_$index, end$\_$index, data$\_$points)\\
\indentintabular Draw a line connecting start point and ending point\\
\indentintabular Calculate the maximum distance each point that lies between start$\_$index and end$\_$index to this line\\
\indentintabular If the maximum distance is greater than a predefined threshold\\
\indentintabular \indentintabular Record the index corresponding to the point that gives the maximum distance\\\\
\indentintabular \indentintabular Set end$\_$index = the index of that point that gives the maximum distance\\
\indentintabular \indentintabular Scan$\_$Line$\_$Approximation (start$\_$index, end$\_$index, data$\_$points)\\\\
\indentintabular \indentintabular Set start$\_$index = the index of that point that gives the maximum distance\\
\indentintabular \indentintabular Scan$\_$Line$\_$Approximation (start$\_$index, end$\_$index, data$\_$points)\\
\indentintabular End if
\\\hline
\end {tabular}}
\end{table}

\begin{figure}[htb]
\centerline{\epsfxsize=4in \epsffile{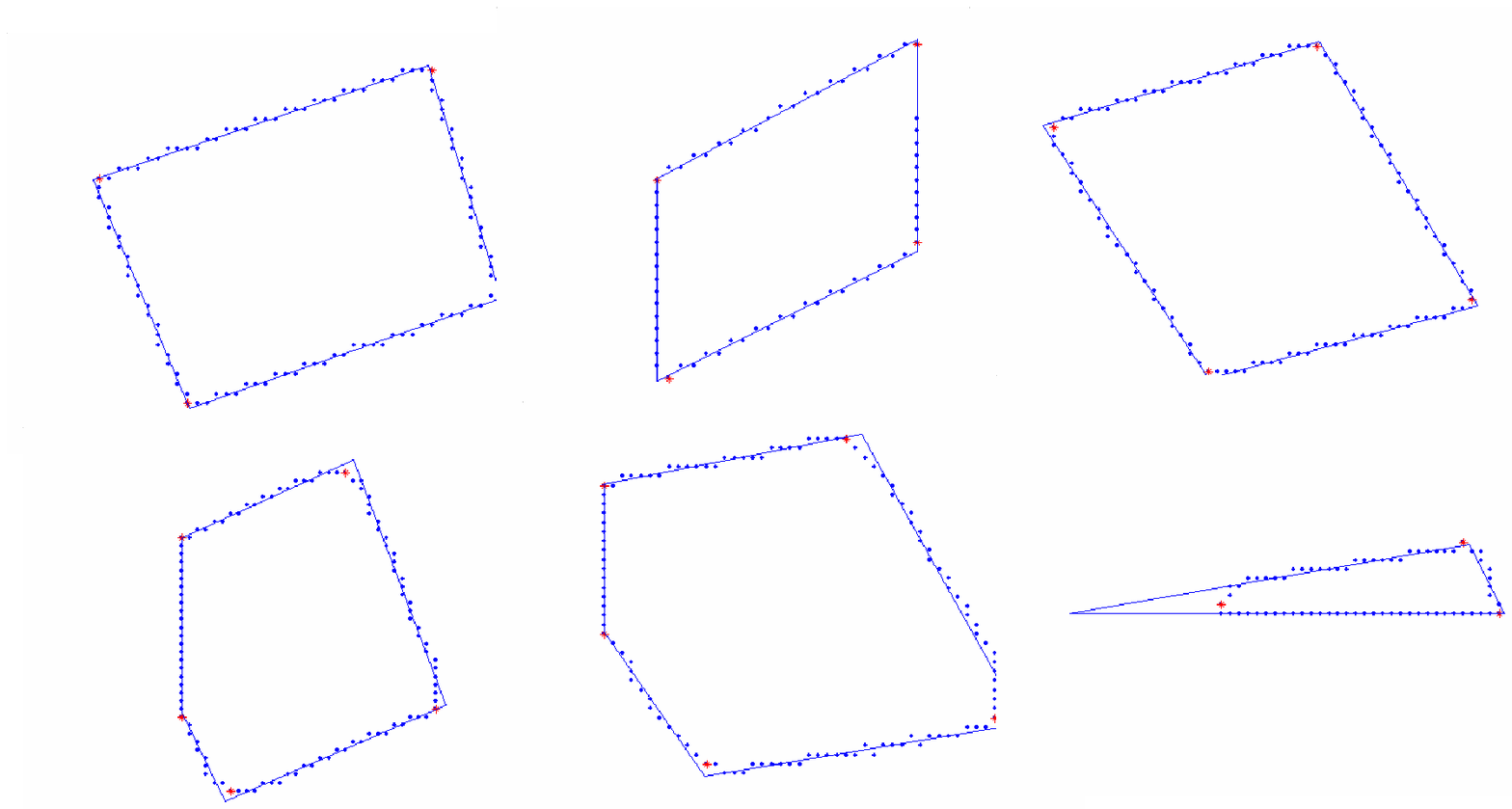}}
\caption {Fitting results using partitions found by scan line approximation algorithm}
\label{fig: scanlineresults}
\end{figure}

In Table \ref{table: scan line approximation algorithm}, the algorithm is described/implemented in a recursive way. Applying this algorithm, an important issue is how to decide the threshold. Unfortunately, this threshold is application-related. In our implementation, we choose $5-10$ pixels. The smaller the boxes or the farther that the camera is way from the calibration object, the smaller the threshold should be. Figure \ref{fig: scanlineresults} shows the fitting results using the partitions found by scan line approximation algorithm, where all the input data are the edge points of some classes in the region map. Another thing that we need to pay attention to is about how to choose the initial starting and ending points. It is obvious that they cannot be on the same side. Otherwise, due to noise in the data, the point whose distance to the line connecting starting and ending points is maximal might not be around corners. That is why we always start around corners as in Figure \ref{fig: scanlineapproximation}. This problem can be solved simply by first finding the two adjacent points whose maximal distance that all other points to this line is the biggest.

\begin{figure}[h]
\centerline{\epsfxsize=5in \epsffile{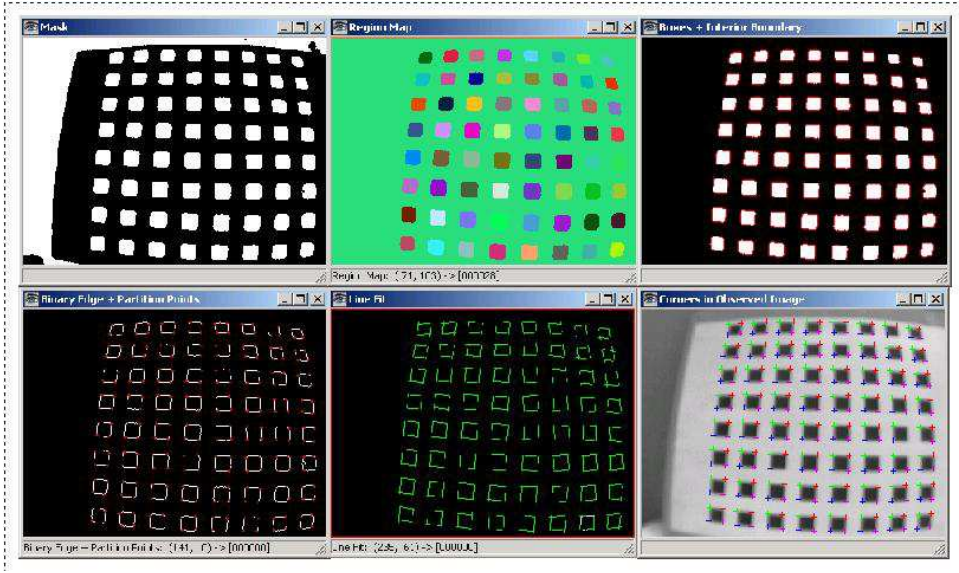}}
\caption {Feature points extraction for desktop images (1)}
\label{fig: feature points extraction for desktop images 1}
\end{figure}

Figure \ref{fig: feature points extraction for desktop images 1}shows an example of the processed images at all steps, where the input images are captured by a desktop camera. Notice that in Figure \ref{fig: feature points extraction for desktop images 2}, in the image titled with ``Binary Image + Partition Points'', the triangular in the upper right corner does not show in the next step. The reason why this happens is that after the process of finding partition points, the number of partition points does not equal to 4 and we thus bypass this class. 

\section{Calibration Method}
\label{section: calibration method}
In this section, the calibration method in \cite{zhang99calibrationinpaper,zhang99calibrationinreport} is described in detail. Using the calibration object shown in Figure \ref{fig: calibration object}, this algorithm is a self-calibration method. It only requires the camera to observe a planar pattern at a few different orientations. Either the camera or the calibration object can be moved by hand and the motion need not be known. The reason why this is feasible is that one image observed by a camera can provide 2 constraints about this camera's intrinsic parameters that are regarded to be unchanged here. With 3 images observed by the same camera, 6 constraints are established and we are able to recover the 5 intrinsic parameters. Once the intrinsic parameters are known, we can estimate the extrinsic parameters, the distortion coefficients $(k_1, k_2)$, and put every initial guess of these parameters into some nonlinear optimization routine to get the final estimations. Another aspect that makes \cite{zhang99calibrationinpaper,zhang99calibrationinreport} appealing is that the author provides calibration results and an executable file on the web page \cite{zhang98calibrationwebpage} along with the sample images. The procedures to do calibration are illustrated in Table \ref{table: camera calibration procedures}. 

\begin{table}[here]
\centering
\caption{Camera Calibration Procedures}
\label{table: camera calibration procedures}
\bigskip
{\begin {tabular}{||l||}\hline\\
\centerline{\bf{Linear Parameter Estimation}}\\
{\bf Estimate Homographies} (Section \ref{section: homography})\\
Let $N$ be the number of images that we want to observe\\
Loop for $i$ from 1 to $N$\\
\indentintabular Assume the calibration object is at $Z^w = 0$\\
\indentintabular Establishes the $i^{th}$  homography between the calibration object and its image \\
\indentintabular Change the orientation of either calibration object or camera\\
End Loop\\\\
{\bf Estimate Intrinsic Parameters} (Section \ref{section: estimation of intrinsic parameters})\\
For each homography we have 2 constraints concerning the 5 intrinsic parameters\\
Now we have $2N$ constraints and we can solve the 5 intrinsic parameters using {\tt SVD}\\\\
{\bf Estimate Extrinsic Parameters} (Section \ref{section: estimation of extrinsic parameters})\\
Using the estimated intrinsic parameters and homographies, we can estimate the extrinsic parameters\\\\
{\bf Estimate Distortion Coefficients} (Section \ref{section: estimation of distortion coeffs})\\
Using the estimated intrinsic and extrinsic parameters, we can get the ideal projected image points\\
Along with the real observed image points, we can estimate the two distortion coefficients $(k_1, k_2)$\\\\
\centerline{\bf{Nonlinear Optimization}}\\
\centerline{(Section \ref{section: nonlinear optimization})}\\
Take all parameters estimated above as an initial guess\\
Use some nonlinear optimization routine, we can get the final estimated values \\
\\\hline
\end {tabular}}
\end{table}

The idea to assume the calibration object is always at $Z^w = 0$ even after some unknown movement maybe bewildering (We are talking about the case when we keep the camera static and move the calibration object). The common sense about the world reference frame is that it is unique. So, how can we assume the calibration object is still at $Z^w = 0$ after some rotation and translation? The answer to this question is: as mentioned before, only the relative position and orientation between the calibration object and the camera is of concern. Each image can provide 2 constraints that are independent to all others. Thinking in the other way, it is the same when we keep the calibration object static and move the camera. The basic calibration equations are given as follows. 

\subsection{Homography Between the Model Plane and Its Image}
\label{section: homography}
Without loss of generality, we assume the calibration object is $Z^w = 0$ in the world reference frame. Let's denote the $i^{th}$  column of the rotation matrix $\bf R$ by $\bf r_i$, we have:
\begin{equation}
\lambda \left [\matrix{
u \cr
v \cr
1} \right ] = {\bf M} \left [\matrix{
X^w \cr
Y^w \cr
Z^w \cr
1} \right ] = {\bf A} \left [\matrix{{\bf r_1} & {\bf r_2}& {\bf r_3} & {\bf t}}\right] \left [\matrix{
X^w \cr
Y^w \cr
Z^w \cr
1} \right ] = {\bf A} \left [\matrix{{\bf r_1} & {\bf r_2}& {\bf t}}\right] \left [\matrix{
X^w \cr
Y^w \cr
1} \right ].
\end{equation}
Therefore a model points in 3-D space is related to its image by a homography $\bf H$
\begin{equation}
\lambda \left[ \matrix {
u \cr
v \cr
1
}\right] = {\bf H} \left [\matrix{
X^w \cr
Y^w \cr
1}\right],
\end{equation}
where \begin{eqnarray}
{\bf H} = {\bf A} \ [{\bf r_1} & {\bf r_2} & {\bf t}]. \nonumber 
\end{eqnarray}
In this way, the $3 \times 3$ matrix $\bf H$ is defined up to a scaling factor.

Given an image of the calibration object, the homography $\bf H$ can be estimated by maximum likelihood criterion. Let $M_i$ and $m_i$ be the model and its image point respectively. Let's assume $m_i$ is corrupted by Gaussian noise with mean 0 and covariance matrix ${\bf \Lambda}_{m_i}$. Then the maximum likelihood estimation of $\bf H$ is obtained by minimizing
\begin {equation}
\sum_i (m_i - {{\hat m}_i})^T {\bf \Lambda}_{m_i} (m_i - {\hat m}_i),
\end{equation}
where ${\bar {\bf h}}_i $ is the $i^{th}$ row of $\bf H$ and 
\begin{equation}
{\hat m}_i = \frac {1}{{{\bar {\bf h}}_3}^T M_i} 
\left[ \matrix{
{\bar {\bf h}}_1^T M_i \cr
{\bar {\bf h}}_2^T M_i
}\right ].
\end {equation}
In practice, we simply assume ${\bf \Lambda}_{m_i} = \sigma ^2 \ {\bf I}$ for all $i$. This is reasonable if the points are extracted independently with the same procedure. For each pair of $M_i$ and $m_i$, we have
\begin{eqnarray}
u &=& \frac{h_{11}X^w+h_{12}Y^w+h_{13}} {h_{31}X^w+h_{32}Y^w+h_{33}} \nonumber \\
v &=& \frac{h_{21}X^w+h_{22}Y^w+h_{23}} {h_{31}X^w+h_{32}Y^w+h_{33}}.
\end{eqnarray}
Let ${\bf x} = [{\bar {\bf h}}_1^T, {\bar {\bf h}}_2^T, {\bar {\bf h}}_3^T]$, then
\begin{equation}
\left [\matrix {
X^w & Y^w & 1 & 0 & 0 & 0 & -uX^w & -uY^w & -u \cr
0 & 0 & 0 & X^w & Y^w & 1 & -vX^w &-vY^w  & -v}\right]{\bf x} = 0.
\end{equation}
When we are given $n$ points, we have $n$ above equations and we can write them in matrix form as $\bf Lx = 0$ where
\[ {\bf L} = \left [ 
\begin{array}{ccccccccc}
X_1^w & Y_1^w & 1 & 0 & 0 & 0 & -u_1X_1^w &-u_1Y_1^w & -u_1 \\
0 & 0 & 0 & X_1^w & Y_1^w & 1 & -v_1X_1^w &-v_1Y_1^w & -v_1 \\
X_2^w & Y_2^w & 1 & 0 & 0 & 0 & -u_2X_2^w &-u_2Y_2^w & -u_2 \\
0 & 0 & 0 & X_2^w & Y_2^w & 1 & -v_2X_2^w &-v_2Y_2^w & -v_2 \\
\multicolumn{9}{c}{\dotfill}\\
X_n^w & Y_n^w & 1 & 0 & 0 & 0 & -u_nX_n^w &-u_nY_n^w & -u_n \\
0 & 0 & 0 & X_n^w & Y_n^w & 1 & -v_nX_n^w &-v_nY_n^w & -v_n 
\end{array} \right ].\]
The matrix $\bf L$ is a $2n\times 9$ matrix and the solution is well known to be the right singular vector of $\bf L$ associated with the smallest singular value.

\subsection{Constraints on the Intrinsic Parameters}
\label{section: intrinsic constraints}
Given the estimated homography ${\bf H} = [{\bf h}_1, {\bf h}_2, {\bf h}_3]$, we have
\begin{equation}
[{\bf h}_1, {\bf h}_2, {\bf h}_3] = \lambda \ {\bf A} \ [{\bf r}_1, {\bf r}_2, {\bf t}],
\end{equation}
with $\lambda$ an arbitrary scalar. Using the knowledge that ${\bf r}_1, {\bf r}_2$ are orthogonal, we have
\begin{eqnarray}
{{\bf r}_1}^T{\bf r}_2 &=& 0 \nonumber \\
{{\bf r}_1}^T{\bf r}_1 &=& {{\bf r}_2}^T{\bf r}_2.
\end{eqnarray}
Since $\lambda {\bf r}_1 = {\bf A}^{-1}{\bf h}_1, \lambda {\bf r}_2 = {\bf A}^{-1}{\bf h}_2$, 
\begin{equation}
\label{eqn: 2 constraints}
\begin{array}{l}
{\bf h}_1^T {\bf A}^{-T}{\bf A}^{-1}{\bf h}_2 = 0 \\
{\bf h}_1^T {\bf A}^{-T}{\bf A}^{-1}{\bf h}_1 = {\bf h}_2^T {\bf A}^{-T}{\bf A}^{-1}{\bf h}_2.
\end{array}
\end{equation}
Given a homography, these are the 2 constraints we obtained on the intrinsic parameters.

\subsection{Estimation of Intrinsic Parameters}
\label{section: estimation of intrinsic parameters}
Let 
\begin{eqnarray}
{\bf B} 
&=& {\bf A}^{-T} {\bf A}^{-1} = 
\left[\matrix{
B_{11} & B_{12} & B_{13} \cr
B_{21} & B_{22} & B_{23} \cr
B_{31} & B_{32} & B_{33}
}\right] \nonumber \\
&=&
{\large
\left [\matrix{
\frac{1}{\alpha^2} & -\frac{\gamma}{\alpha^2\beta}& \frac{v_0\gamma - u_0 \beta}{\alpha^2\beta} \cr
-\frac{\gamma}{\alpha^2 \beta} & \frac{\gamma^2}{\alpha^2 \beta^2} + \frac{1}{\beta^2} & -\frac{\gamma(v_0 \gamma - u_0 \beta)}{\alpha^2\beta^2} - \frac{v_0}{\beta^2} \cr
\frac{v_0\gamma - u_0\beta}{\alpha^2\beta} & -\frac{\gamma(v_0 \gamma - u_0 \beta)}{\alpha^2\beta^2} - \frac{v_0}{\beta^2} & 
\frac{(v_0\gamma - u_0\beta)^2}{\alpha^2\beta^2} + \frac{v_0^2}{\beta^2} + 1
}\right]}.
\end{eqnarray}
Note that $\bf B$ is symmetric. Define ${\bf b} = [B_{11}, B_{12}, B_{22}, B_{13}, B_{23}, B_{33}]^T$. Let the $i^{th}$ column vector of $\bf H$ be ${\bf h}_i = [h_{i1}, h_{i2}, h_{i3}]^T$, we have

\begin{eqnarray}
{\bf h}_i^T{\bf B}{\bf h}_j &=&  [h_{i1} \ h_{i2} \ h_{i3}]
\left[ \matrix{
B_{11} & B_{12} & B_{13} \cr
B_{21} & B_{22} & B_{23} \cr
B_{31} & B_{32} & B_{33}} \right]
\left[\matrix{
h_{j1} \cr
h_{j2} \cr
h_{j3}}\right] \nonumber \\
&=& h_{j1}(h_{i1}B_{11}+h_{i2}B_{12}+h_{i3}B_{13}) + h_{j2}(h_{i1}B_{12}+h_{i2}B_{22}+h_{i3}B_{23}) \nonumber \\
&+& h_{j3}(h_{i1}B_{13} +h_{i2}B_{23}+h_{i3}B_{33}).
\end{eqnarray}
Denote \begin{eqnarray}
{\bf V}_{ij}= [h_{i1}h_{j1}, \ h_{i1}h_{j2}+h_{i2}h_{j1}, \ h_{i2}h_{j2}, \ h_{i3}h_{j1}+h_{i1}h_{j3}, \ h_{i3}h_{j2}+h_{i2}h_{j3}, \ h_{i3}h_{j3}]^T,
\end{eqnarray} the two constraints in Equation (\ref{eqn: 2 constraints}) become
\begin{eqnarray}
\left[\matrix{
{\bf V}_{12}^T \cr
({\bf V}_{11}-{\bf V}_{12})^T
}\right] {\bf b} = {\bf 0}.
\end{eqnarray}
If $N$ images of the calibration object are taken, by stacking $N$ such equations, we have $ {\bf V}\bf b = \bf 0$ where $\bf V$ is a $2N\times 6$ matrix. 

When $N \ge 3$, we will have a unique solution defined up to a scaling factor. The solution is well known to be the right singular vector of $\bf V$ associated with the smallest singular value. The matrix $\bf B$ is estimated up to a scaling factor ${\bf B}=\lambda \ {\bf A}^{-T}{\bf A}^{-1}$. After estimation of $\bf B$, the intrinsic parameters can be extracted from $\bf B$ by
\begin{eqnarray}
\label{eqn: intrinsic parameters from B}
v_0     & = & (B_{12}B_{13} - B_{11}B_{23}) / (B_{11}B_{22}-B_{12}^2) , \nonumber \\
\lambda & = & B_{33} - [B_{13}^2 + v_0(B_{12}B_{13} -B_{11}B_{23})] /B_{11} , \nonumber \\
\alpha  & = & \sqrt{\lambda / B_{11}} , \\
\beta   & = & \sqrt{\lambda B_{11} /(B_{11}B_{22}-B_{12}^2)} , \nonumber \\
\gamma  & = & -B_{12}\alpha^2\beta/\lambda , \nonumber \\
u_0     & = & \gamma  v_0 /\beta - B_{13}\alpha^2/\lambda . \nonumber
\end{eqnarray}

The original equation to estimate $u_0$ in \cite{zhang99calibrationinpaper,zhang99calibrationinreport} is $u_0 = \gamma v_0/\alpha - B_{13}\alpha^2/\lambda$. This must be an obvious mistake since when all the other 5 parameters are known, $u_0$ can be estimated directly from $B_{13}$. The reason why using a wrong equation to estimate $u_0$ still achieves a good accuracy might due to the fact that $\alpha$ and $\beta$ are the scaling factors in the two image axes and they are usually close to each other.

\subsection{Estimation of Extrinsic Parameters}
\label{section: estimation of extrinsic parameters}
Once $\bf A$ is known, the extrinsic parameters can be estimated as:
\begin{eqnarray}
{\bf r}_1 &=& \lambda \ {\bf A}^{-1}{\bf h}_1 	\nonumber \\
{\bf r}_2 &=& \lambda \ {\bf A}^{-1}{\bf h}_2 	\\
{\bf r}_3 &=& {\bf r}_1 \times {\bf r}_2 		\nonumber \\
{\bf t}   &=& \lambda \ {\bf A}^{-1}{\bf h}_3 	\nonumber 
\end{eqnarray}
where $\lambda = 1/||{\bf A}^{-1}{\bf h}_1||_2 = 1/||{\bf A}^{-1}{\bf h}_2||_2$.
Of course, due to the noise in the data, the computed matrix ${\bf R} = [{\bf r}_1 \ {\bf r}_2 \ {\bf r}_3]$ does not satisfy the properties of a rotation matrix ${\bf R}{\bf R}^T = {\bf I}$. One way to estimate the best rotation matrix from a general $3\times3$ matrix $\bf R$ is:   by the Matlab function {\tt svd} with $[{\bf U},{\bf S},{\bf V}] = {\tt svd} \;({\bf R})$. The best rotation matrix will be ${\bf UV}^T$. 

\subsection{Estimation of Distortion Coefficients}
\label{section: estimation of distortion coeffs}
Assume the center of distortion is the same as the principal point, Equation (\ref{eqn: radial distortion}) describes the relationship between the ideal projected undistorted image points $(u,v)$ and the real observed distorted image points $(u_d, v_d)$. Given $n$ points in $N$ images, we can stack all equations together to obtain totally $2Nn$ equations in matrix form as $\bf D\bf k=\bf d$, where ${\bf k}=[k_1, k_2]^T$. The linear least-square solutions for $\bf k$ is
\begin{equation}
{\bf k} = ({\bf D}^T{\bf D})^{-1}{\bf D}^T{\bf d}.
\end{equation}

\subsection{Nonlinear Optimization: Complete Maximum Likelihood Estimation}
\label{section: nonlinear optimization}
Assume that the image points are corrupted independently by identically distributed noise, the maximum likelihood estimation can be obtained by minimizing the following objective function
\begin{equation}
\label{eqn: objective function}
J = \sum_{i=1}^N \sum_{j=1}^n ||m_{ij}-\hat m({\bf A},k_1, k_2, {\bf R}_i, {\bf t}_i, M_j)||^2,
\end{equation}
where $\hat m({\bf A},k_1, k_2, {\bf R}_i, {\bf t}_i, M_j)$ is the projection of point $M_j$ in the $i^{th}$image using the estimated parameters. This is a nonlinear optimization problem that can be solved by Matlab optimization function {\tt fminunc}.

In our implementation, one observation is that without an initial estimation of distortion coefficients $(k_1, k_2)$, simply setting them to be 0, gives the same optimization results as the case with a good initial guess. Clearly, a ``good'' initial guess of the distortion coefficients is not practically required.

\section{Calibration of Different Cameras}
In this section, some calibration results are presented using the images provided in \cite{zhang98calibrationwebpage}. Images captured by a desktop camera and the ODIS camera are also used. For each camera, $5$ images are captured and the feature locations are extracted for each image. Here, we always use 5 images for calibration. Using different number of images is also feasible. In practice, we found that 5 images are sufficient for camera calibration.

\subsection{Code Validation Using Images in \cite{zhang99calibrationinpaper,zhang99calibrationinreport}}
In \cite{zhang99calibrationinpaper,zhang99calibrationinreport}, the calibration images are posted on web page \cite{zhang98calibrationwebpage}. We use the reported results to validate our implementation code with the same calibration images. 

\subsubsection{Plot of the Observed and the Projected Image Points - Microsoft Images}
\begin{figure}[htb]
\begin{minipage}{\textwidth}
\setcounter{mpfootnote}{\value{footnote}}
\renewcommand{\thempfootnote}{\arabic{mpfootnote}}
\centerline{\epsfxsize=4in \epsffile{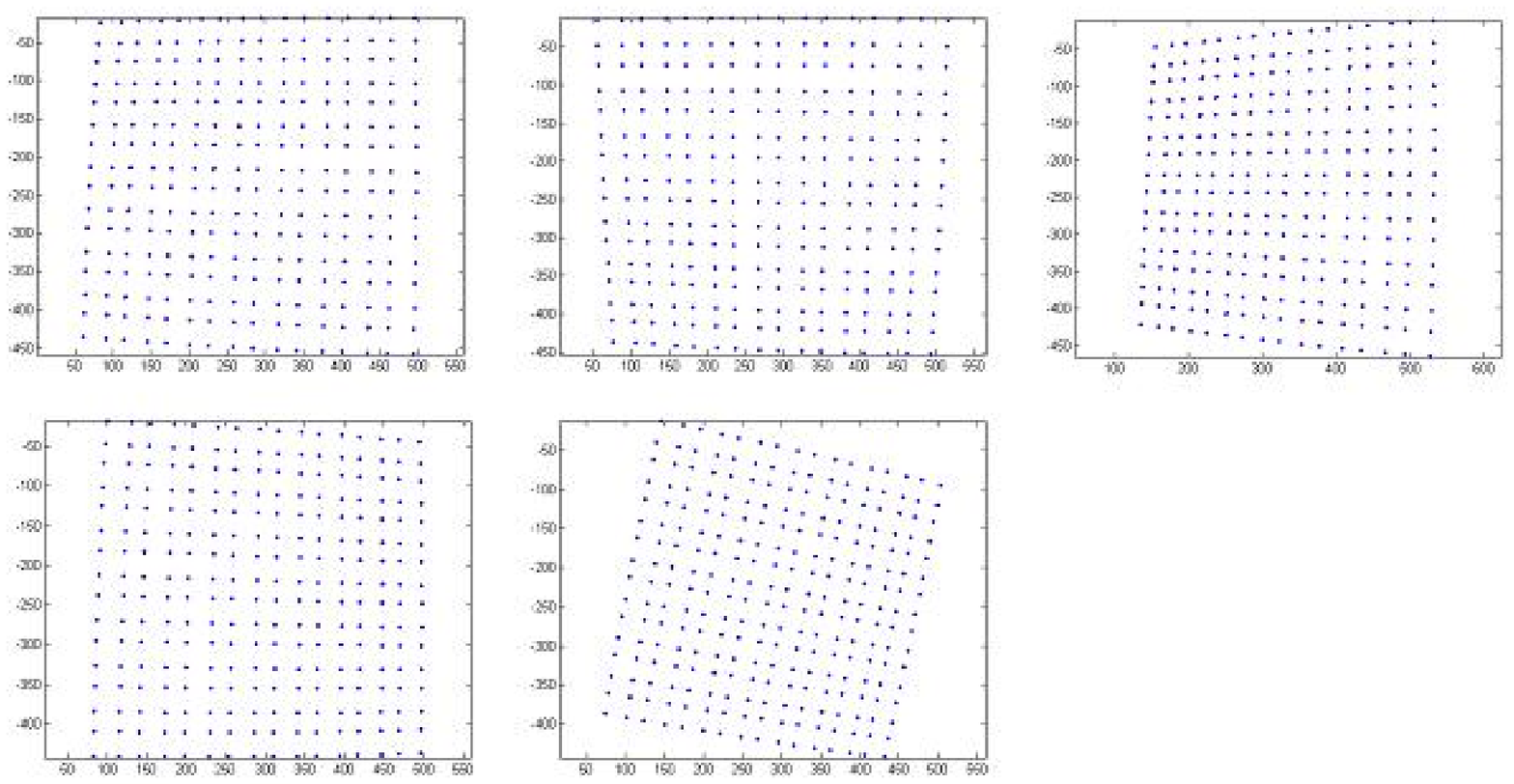}}
\caption[Plot of the observed and the projected image points - Microsoft Images]%
{Plot of the observed and the projected image points - Microsoft images \footnote{Blue dots are the real observed image points}\footnote{Red dots are the projected image points using the estimated camera parameters}}
\label{fig: calibration_results_plot_ms}
\end{minipage}
\end{figure}

Figure \ref{fig: calibration_results_plot_ms} shows the observed and the projected image points using the images provided in \cite{zhang99calibrationinpaper,zhang99calibrationinreport}.

\subsubsection {Comparison of Calibration Results - Microsoft Images}
The parameters before and after nonlinear optimization along with the Microsoft calibration results, obtained by executing its executable file posted on the web page \cite{zhang98calibrationwebpage}, are shown in Table \ref{table: Comparison of Calibration Results for Microsoft Images}. Comparing these final calibration results, one can find that there are slight differences between some of the parameters such as $v_0$. However, as can be seen in section \ref{section: objective function - Microsoft images}, the objective functions from these two calibration codes are very close. 

{\small                                                                       
\begin{table}[htb]                                                                 
\centering                                                                         
\caption{Comparison of Calibration Results - Microsoft Images}
\label{table: Comparison of Calibration Results for Microsoft Images}              
\renewcommand{\arraystretch}{0.92}                                                 
\bigskip                                                                           
{\begin {tabular}{|r|r|r|r|}\hline                                                                        
& \multicolumn{2}{|c|}{\bf Our Implementation} & {\bf Microsoft} \\\hline                       
& \multicolumn{1}{|r|}{\bf Before Opti} & {\bf After Opti} & {\bf After Opti} \\\hline
$\alpha$ & 871.4450 &   832.5010 &   832.5     \\\hline                                  
$\gamma$ & 0.2419   &     0.2046 &   0.2045    \\\hline                                  
$u_0$    & 300.7676 &   303.9584 &   303.959   \\\hline                                  
$\beta$  & 871.1251 &   832.5309 &   832.53    \\\hline                                  
$v_0$    & 220.8684 &   206.5879 &   206.585   \\\hline                                  
$k_1$    & 0.1371   &    -0.2286 &   -0.2286   \\\hline                                  
$k_2$    & -2.0101  &     0.1903 &   0.1903    \\\hline
\end {tabular}}                                                                    
\end{table}}

\subsubsection{Objective Function - Microsoft Images}
\label{section: objective function - Microsoft images}
Table \ref{table: objective function - ms} shows the comparison of the final values of objective function defined in Equation (\ref{eqn: objective function}) after nonlinear optimization between Microsoft result and our implementation. The results show that they are very close. We can conclude that our code is correct for the Microsoft images. In what follows, we shall present two more groups of calibration results for our desktop camera and the ODIS camera in our center. As in the case of Microsoft images, we will compare the results similarly for further validation. \begin{table}[h]
\centering
\caption{Objective Function - Microsoft Images}
\label{table: objective function - ms}
\bigskip
	{\begin {tabular}{|c|c|}\hline
	\bf Microsoft & \bf Our Code\\
	\hline
	144.8799 & 144.8802
	\\\hline
	\end {tabular}}
\end{table}

\begin{remark}
The options when using the Matlab function {\tt fminunc} is not recorded. So, when using different options, slightly different results can be achieved. 
\end{remark}

\subsubsection{Nonlinear Optimization Iterations - Microsoft Images}
Table \ref{table: Nonlinear Optimization Iterations - Microsoft Images} shows the nonlinear optimization iterations, where the initial guess of all parameters are the estimations obtained in Section \ref{section: calibration method}.  The nonlinear optimization using Matlab function {\tt fminunc}. From this table, we can see that after 52 iterations, the value of $f(x)$, the objective function defined in Equation (\ref{eqn: objective function}), drops to $144.88$ from the initial value of $1055.89$. 
{\footnotesize
\begin{table}[htb]
\centering
\caption{Nonlinear Optimization Iterations - Microsoft Images}
\label{table: Nonlinear Optimization Iterations - Microsoft Images}
\renewcommand{\arraystretch}{0.9}
\bigskip
	{\begin {tabular}{|r|r|r|r|r|}\hline
	{\bf Iteration} & {\bf Function} & $f(x)$ & {\bf Step-size} & {\bf Directional} \\
	& {\bf Count} & & &{\bf Derivative} \\\hline
   1   &     37    &      1055.89  &          0.001   &   -5.05e+009   \\
   2   &     78    &      1032.26  &   9.36421e-009   &   -7.97e+004   \\
   3   &    120    &      915.579  &   1.85567e-007   &   -1.13e+004   \\
   4   &    161    &       863.19  &    1.1597e-008   &   -2.78e+005   \\
   5   &    202    &      860.131  &   1.77145e-008   &   -2.45e+004   \\
   6   &    244    &      836.386  &   1.37495e-007   &   -4.83e+003   \\
   7   &    285    &      820.765  &   1.58388e-008   &   -1.13e+005   \\
   8   &    327    &       816.13  &   1.86391e-007   &      -2e+003   \\
   9   &    368    &      800.842  &   3.79421e-008   &   -4.12e+004   \\
  10   &    410    &      788.888  &   3.68321e-007   &   -3.29e+003   \\
  11   &    452    &      769.459  &   1.47794e-006   &   -1.38e+003   \\
  12   &    493    &      738.541  &   1.13935e-006   &     3.2e+003   \\
  13   &    535    &      692.716  &   3.81991e-006   &          606   \\
  14   &    576    &      674.548  &    1.0489e-006   &          360   \\
  15   &    618    &      631.838  &   3.81559e-006   &   -1.26e+003   \\
  16   &    659    &      616.973  &   1.91963e-006   &   -1.79e+003   \\
  17   &    700    &      604.857  &   2.96745e-006   &         -724   \\
  18   &    742    &      593.179  &   6.19011e-006   &         -980   \\
  19   &    783    &       573.66  &   3.68278e-006   &   -4.67e+003   \\
  20   &    824    &      544.497  &   4.69657e-006   &   -2.28e+003   \\
  21   &    865    &      537.636  &   6.60037e-006   &        -15.1   \\
  22   &    906    &      530.468  &   3.62979e-006   &         -103   \\
  23   &    947    &      525.032  &   2.35736e-006   &        -85.4   \\
  24   &    989    &      523.091  &   4.80959e-006   &         22.9   \\
  25   &   1031    &      509.698  &   2.98894e-005   &         -320   \\
  26   &   1072    &      505.972  &    2.3338e-006   &   -2.54e+003   \\
  27   &   1114    &      499.005  &    6.5114e-005   &        -61.8   \\
  28   &   1155    &      493.817  &   4.78348e-006   &         -587   \\
  29   &   1197    &      468.823  &    0.000433333   &         -162   \\
  30   &   1238    &      378.238  &    0.000463618   &         -924   \\
  31   &   1279    &       280.68  &     0.00024013   &   -1.45e+003   \\
  32   &   1320    &      230.465  &    0.000148949   &         -990   \\
  33   &   1361    &      223.328  &     0.00017819   &         31.5   \\
  34   &   1402    &      221.525  &      0.0649328   &         10.6   \\
  35   &   1444    &      218.413  &       0.179589   &    -0.000716   \\
  36   &   1486    &      191.642  &       0.577805   &       -0.521   \\
  37   &   1528    &      176.735  &        1.68909   &        0.003   \\
  38   &   1570    &      173.092  &        1.62593   &    8.26e-005   \\
  39   &   1612    &      169.211  &        1.38363   &      -0.0033   \\
  40   &   1654    &      157.984  &         2.9687   &      0.00291   \\
  41   &   1696    &      146.471  &        1.87697   &      -0.0366   \\
  42   &   1737    &      145.015  &       0.894999   &      0.00491   \\
  43   &   1778    &      144.911  &       0.869282   &    -0.000345   \\
  44   &   1820    &      144.893  &        1.56986   &     0.000205   \\
  45   &   1862    &      144.882  &        1.50158   &    5.98e-005   \\
  46   &   1903    &       144.88  &        1.31445   &     -0.00126   \\
  47   &   1946    &       144.88  &      0.0356798   &     0.000261   \\
  48   &   1989    &       144.88  &        1.06404   &   -9.25e-006   \\
  49   &   2030    &       144.88  &       0.771663   &    1.23e-005   \\
  50   &   2071    &       144.88  &       0.123009   &    -0.000482   \\
  51   &   2109    &       144.88  &      0.0615043   &    -0.000114   \\
  52   &   2147    &       144.88  &     -0.0307522   &   -1.18e-005   \\\hline
	\end {tabular}}
\end{table}}

\clearpage

\subsection{Calibration of a Desktop Camera}
This section shows the calibration results for a desktop camera. 

\subsubsection{Extracted Corners in the Observed Images - The Desktop Camera Case}
\begin{figure}[htb]
\centerline{\epsfxsize=7in \epsffile{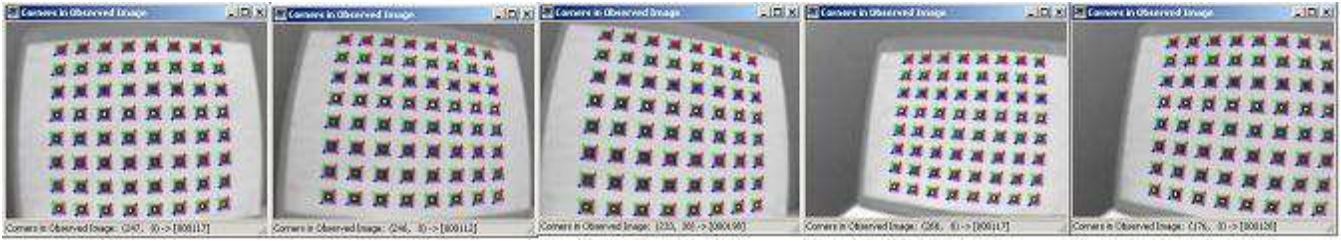}}
\caption {Extracted corners in the observed images captured by the desktop camera}
\label{fig: extracted corners desktop}
\end{figure}

Figure \ref{fig: extracted corners desktop} shows the extracted feature points in the observed images captured by the desktop camera. The extracted corners are marked by a cross and the dot in the center of each box is just to test if the detected boxes are in the same order as the 3-D points in the world reference frame. Due to the low accuracy of this camera, the extracted feature points deviate a lot from their ``true positions'', as ``sensed'' or ``perceived'' by our human observers. 

\subsubsection{Plot of the Observed and the Projected Image Points - The Desktop Camera Case}
\begin{figure}[htb]
\centerline{\epsfxsize=4.8in \epsffile{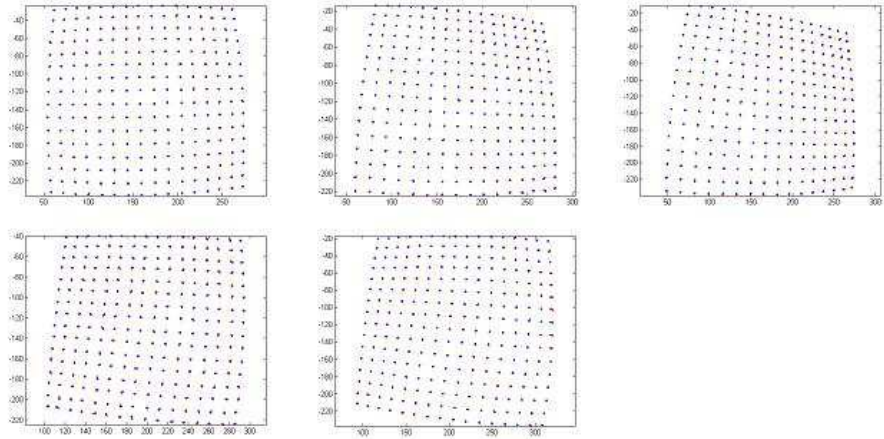}}
\caption {Plot of the observed and the projected image points - the desktop camera case}
\label{fig: calibration_results_plot_desktop}
\end{figure}

Figure \ref{fig: calibration_results_plot_desktop} shows the observed and the projected image points captured by the desktop camera. For descriptions, please refer to Figure \ref{fig: calibration_results_plot_ms}. 

\subsubsection {Comparison of Calibration Results - The Desktop Camera Case}
Table \ref{table: Comparison of Calibration Results for Desktop Images} show the calibration results of our implementation and Microsoft executable file. 
{\small                                                                       
\begin{table}[htb]                                                                 
\centering                                                                         
\caption{Comparison of Calibration Results - The Desktop Camera Case}
\label{table: Comparison of Calibration Results for Desktop Images}              
\renewcommand{\arraystretch}{1}                                                 
\bigskip                                                                           
{\begin {tabular}{|r|r|r|r|}\hline                                                                        
& \multicolumn{2}{|c|}{\bf Our Implementation} & {\bf Microsoft} \\\hline                       
& \multicolumn{1}{|r|}{\bf Before Opti} & {\bf After Opti} & {\bf After Opti} \\\hline
$\alpha$ & 350.066701&  277.1457   & 277.145  \\\hline
$\gamma$ & 1.693062  &     -0.5730 &  -0.573223 \\\hline
$u_0$    & 200.051398&    153.9923 & 153.989  \\\hline
$\beta$  & 342.500985&    270.5592 & 270.558  \\\hline
$v_0$    & 100.396596&    119.8090 &  119.812   \\\hline
$k_1$    & 0.096819  &     -0.3435 &  -0.343527 \\\hline
$k_2$    & -0.722239 &      0.1232 & 0.123163   \\\hline
\end {tabular}}                                                                    
\end{table}}

\subsubsection{Objective Function - The Desktop Camera Case}
From Table \ref{table: objective function - desktop}, we can see that the final calibration results by our implementation and by the Microsoft group are almost identical. 
\begin{table}[h]
\centering
\caption{Objective Function - The Desktop Camera Case}
\label{table: objective function - desktop}
\bigskip
	{\begin {tabular}{|c|c|}\hline
	\bf Microsoft & \bf Our Code\\
	\hline
	778.9763 & 778.9768
	\\\hline
	\end {tabular}}
\end{table}

\subsubsection{Nonlinear Optimization Iterations - The Desktop Camera Case}
For data format and descriptions, please refer to Table \ref{table: Nonlinear Optimization Iterations - Microsoft Images}. 
{\scriptsize
\begin{table}[htb]
\centering
\caption{Nonlinear Optimization Iterations - The Desktop Camera Case}
\label{table: Nonlinear Optimization Iterations - desktop Images}
\renewcommand{\arraystretch}{0.89}
\bigskip
	{\begin {tabular}{|r|r|r|r|r|}\hline
	{\bf Iteration} & {\bf Function} & $f(x)$ & {\bf Step-size} & {\bf Directional} \\
	& {\bf Count} & & &{\bf Derivative} \\\hline
     1  &     37  &    7077.09 &          0.001  &     -6.46e+009   \\
     2  &     79  &    6943.03 &   4.15059e-008  &      1.29e+005   \\        
     3  &    121  &    6821.48 &   2.25034e-007  &     -3.08e+003   \\         
     4  &    162  &    6760.09 &   5.93771e-008  &     -2.22e+004   \\         
     5  &    204  &    6747.99 &   1.11643e-007  &     -2.72e+003   \\         
     6  &    246  &    6574.48 &   8.41756e-007  &     -2.97e+004   \\         
     7  &    287  &    6543.07 &   4.72854e-008  &      -2.6e+004   \\         
     8  &    329  &    6428.87 &   6.11914e-007  &     -1.47e+003   \\         
     9  &    370  &    6386.18 &   5.19775e-008  &      -1.8e+004   \\         
    10  &    412  &     6108.9 &   1.57962e-006  &     -1.37e+005   \\         
    11  &    453  &    5961.18 &    2.7321e-006  &           -940   \\         
    12  &    495  &    5886.28 &   1.25318e-005  &     -5.55e+003   \\         
    13  &    536  &    5825.17 &   1.33725e-005  &     -1.96e+003   \\         
    14  &    577  &    5784.52 &   1.59682e-005  &           -710   \\         
    15  &    619  &    5727.74 &   6.05741e-005  &           -192   \\         
    16  &    661  &    5519.12 &    0.000129424  &           -292   \\         
    17  &    702  &    4413.34 &    0.000150092  &     -4.65e+005   \\         
    18  &    743  &    4214.43 &   3.64671e-005  &     -4.58e+004   \\         
    19  &    784  &    4106.31 &   4.97942e-005  &           -426   \\         
    20  &    825  &    4062.22 &   3.43799e-005  &           -546   \\         
    21  &    867  &    3974.48 &    0.000114609  &           -298   \\         
    22  &    908  &    3879.17 &   9.30444e-005  &            413   \\         
    23  &    949  &    3808.83 &   8.52215e-005  &           -158   \\         
    24  &    990  &    3766.27 &   5.46297e-005  &           -865   \\         
    25  &   1032  &    3691.29 &    0.000156467  &           -263   \\         
    26  &   1073  &    3622.83 &    0.000149539  &           -364   \\         
    27  &   1114  &    3607.38 &    0.000119243  &           4.45   \\         
    28  &   1156  &    3591.08 &    0.000379983  &          -6.78   \\         
    29  &   1197  &    3575.83 &    0.000240085  &          -21.2   \\         
    30  &   1239  &       3556 &    0.000489406  &          -2.82   \\         
    31  &   1280  &    3549.92 &    0.000163356  &          -23.3   \\         
    32  &   1322  &    3420.37 &     0.00606838  &          -44.4   \\         
    33  &   1363  &     3285.3 &     0.00265288  &           -305   \\         
    34  &   1404  &     3268.6 &     0.00371299  &          -3.06   \\         
    35  &   1446  &    2911.69 &      0.0421166  &           -284   \\         
    36  &   1487  &    2502.19 &      0.0301324  &           -809   \\         
    37  &   1529  &    1810.42 &       0.841983  &          -2.73   \\         
    38  &   1570  &    1590.51 &       0.863446  &          -1.84   \\         
    39  &   1611  &    1457.62 &       0.317624  &          -10.9   \\         
    40  &   1652  &     1224.1 &       0.681949  &          -73.5   \\         
    41  &   1693  &    1147.71 &       0.316197  &          -32.5   \\         
    42  &   1734  &    1061.67 &       0.479219  &         -0.771   \\         
    43  &   1775  &    990.417 &       0.423682  &        -0.0632   \\         
    44  &   1816  &    944.745 &         0.4477  &         -0.602   \\         
    45  &   1857  &    892.944 &       0.711193  &         -0.207   \\         
    46  &   1898  &    871.868 &       0.552265  &         -0.523   \\         
    47  &   1939  &    855.089 &       0.463739  &         0.0449   \\         
    48  &   1981  &    832.165 &         1.1442  &          -0.16   \\         
    49  &   2022  &    818.818 &       0.517951  &        -0.0946   \\         
    50  &   2063  &    811.471 &       0.697572  &       -0.00811   \\         
    51  &   2104  &    806.001 &        0.58951  &        0.00154   \\         
    52  &   2146  &    799.658 &       0.894743  &       -0.00199   \\         
    53  &   2187  &    794.968 &        0.97607  &        -0.0104   \\         
    54  &   2228  &    789.198 &       0.923713  &       -0.00576   \\         
    55  &   2269  &    786.548 &       0.653634  &       -0.00545   \\         
    56  &   2310  &    783.866 &       0.706569  &       -0.00184   \\         
    57  &   2351  &    781.579 &       0.908965  &      -0.000913   \\         
    58  &   2392  &    780.375 &       0.925928  &       -0.00159   \\         
    59  &   2433  &    779.786 &       0.642352  &      7.13e-005   \\         
    60  &   2474  &    779.452 &       0.783045  &     -9.29e-005   \\         
    61  &   2515  &    779.317 &        1.00864  &      -0.000402   \\         
    62  &   2556  &    779.175 &        1.36328  &        -4e-005   \\         
    63  &   2597  &     779.07 &        1.16283  &       0.000299   \\         
    64  &   2638  &    779.024 &       0.799815  &       0.000659   \\         
    65  &   2680  &    779.004 &        1.26311  &      -0.000271   \\         
    66  &   2722  &    778.987 &        1.75016  &     -1.44e-005   \\         
    67  &   2763  &    778.979 &         1.3062  &     -6.31e-006   \\         
    68  &   2804  &    778.978 &       0.956557  &     -1.84e-005   \\         
    69  &   2846  &    778.977 &        1.36853  &     -2.32e-006   \\         
    70  &   2887  &    778.977 &          1.191  &      1.01e-006   \\         
    71  &   2928  &    778.977 &       0.736495  &     -4.04e-006   \\         
    72  &   2969  &    778.977 &       0.498588  &      -0.000232   \\         
    73  &   3007  &    778.977 &       0.249294  &     -1.46e-005   \\\hline
	\end {tabular}}
\end{table}}

\clearpage

\subsection{Calibration of the ODIS Camera}
Now, we want to calibrate the ODIS camera, a 1/4 inch color board camera with 2.9 mm Lens \cite{odiscamera}. 

\subsubsection{Extracted Corners in the Observed Images - The ODIS Camera Case}
\begin{figure}[htb]
\centerline{\epsfxsize=7in \epsffile{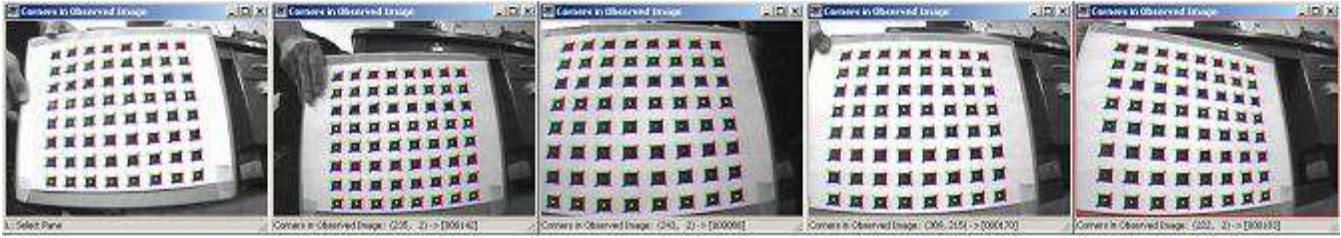}}
\caption {Extracted corners in observed images captured by the ODIS camera}
\label{fig: extracted corners ODIS}
\end{figure}

Figure \ref{fig: extracted corners ODIS} shows the extracted feature points in the observed images captured by the ODIS camera. 

\subsubsection{Plot of the Observed and the Projected Image Points - The ODIS Camera Case}
\begin{figure}[htb]
\centerline{\epsfxsize=5.4in \epsffile{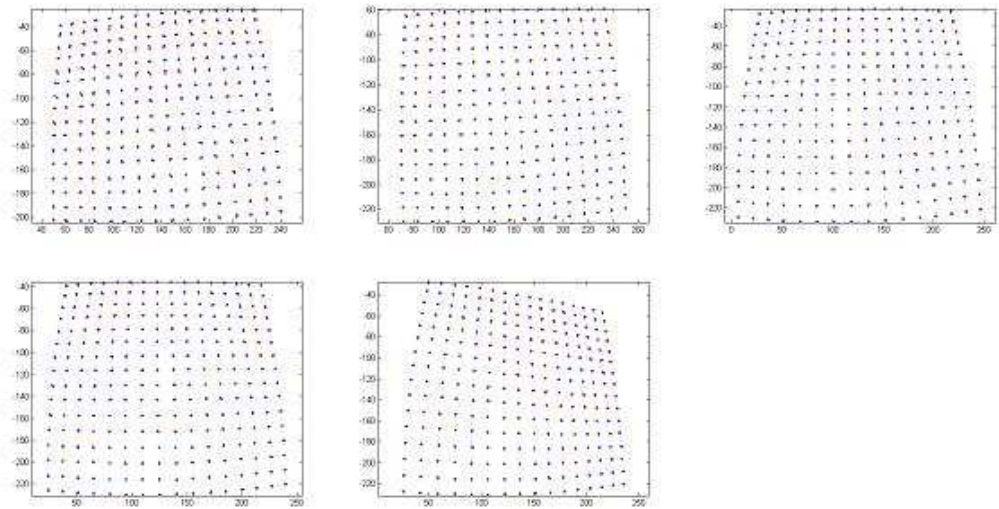}}
\caption {Plot of the observed and the projected image points - the ODIS camera case}
\label{fig: calibration_results_plot_ODIS}
\end{figure}

Figure \ref{fig: calibration_results_plot_ODIS} shows the observed and the projected image points captured by the ODIS camera. For descriptions, please refer to Figure \ref{fig: calibration_results_plot_ms}. 

\subsubsection {Comparison of Calibration Results - The ODIS Camera Case}
(See Table \ref{table: Comparison of Calibration Results for ODIS Images})
{\small                                                                       
\begin{table}[htb]                                                                 
\centering                                                                         
\caption{Comparison of Calibration Results - The ODIS Camera Case}
\label{table: Comparison of Calibration Results for ODIS Images}              
\renewcommand{\arraystretch}{1}                                                 
\bigskip                                                                           
{\begin {tabular}{|r|r|r|r|}\hline                                                                        
& \multicolumn{2}{|c|}{\bf Our Implementation} & {\bf Microsoft} \\\hline                       
& \multicolumn{1}{|r|}{\bf Before Opti} & {\bf After Opti} & {\bf After Opti} \\\hline
$\alpha$ & 320.249458&  260.7636   &  260.764    \\\hline
$\gamma$ & 10.454189 &     -0.2739 &  -0.273923  \\\hline
$u_0$    & 164.735845&    140.0564 &  140.056    \\\hline
$\beta$  & 306.001053&    255.1465 &  255.147    \\\hline
$v_0$    & 85.252209 &    113.1723 &  113.173    \\\hline
$k_1$    & 0.071494  &     -0.3554 &   -0.355429 \\\hline
$k_2$    & -0.342866 &      0.1633 &  0.163272   \\\hline
\end {tabular}}                                                                    
\end{table}}

\subsubsection{Objective Function - The ODIS Camera Case}
(See Table \ref{table: objective function - ODIS})
\begin{table}[here]
\centering
\caption{Objective Function - The ODIS Camera Case}
\label{table: objective function - ODIS}
\bigskip
	{\begin {tabular}{|c|c|}\hline
	\bf Microsoft & \bf Our Code\\
	\hline
	840.2665 & 840.2650
	\\\hline
	\end {tabular}}
\end{table}

\subsubsection{Nonlinear Optimization Iterations - The ODIS Camera Case}
(See Table \ref{table: Nonlinear Optimization Iterations - ODIS Images})
{\scriptsize
\begin{table}[htb]
\centering
\caption{Nonlinear Optimization Iterations - The ODIS Camera Case}
\label{table: Nonlinear Optimization Iterations - ODIS Images}
\renewcommand{\arraystretch}{0.88}
\bigskip
	{\begin {tabular}{|r|r|r|r|r|}\hline
	{\bf Iteration} & {\bf Function} & $f(x)$ & {\bf Step-size} & {\bf Directional} \\
      & {\bf Count} & & &{\bf Derivative} \\\hline
     1     &    37    &     5872.79   &         0.001    & -9.99e+008       \\       
     2     &    78    &      5849.4   &  4.67898e-008    &  -1.93e+005      \\       
     3     &   120    &     5841.17   &  1.15027e-007    &  -2.18e+003      \\       
     4     &   162    &     5734.46   &  7.19555e-007    &  -3.04e+003      \\       
     5     &   203    &     5703.36   &  1.27443e-007    &  -6.05e+003      \\       
     6     &   245    &     5524.51   &  4.61815e-007    &  -1.11e+004      \\       
     7     &   286    &     5505.25   &  7.93522e-008    &  -7.52e+003      \\       
     8     &   328    &     5286.49   &  9.13589e-007    &  -3.24e+004      \\       
     9     &   369    &     5252.79   &  4.27308e-007    &  -1.41e+003      \\       
    10     &   410    &     5147.55   &   2.0105e-007    &  -5.05e+004      \\       
    11     &   452    &     5123.77   &  1.07891e-006    &        -278      \\       
    12     &   494    &     5111.44   &  8.41391e-006    &       -59.2      \\       
    13     &   535    &     5085.05   &  4.14977e-006    &        -301      \\       
    14     &   577    &      5033.1   &    0.00012499    &       -49.6      \\       
    15     &   618    &      5017.1   &  8.89033e-006    &  -4.93e+003      \\       
    16     &   660    &     4875.16   &  3.23114e-005    &   -3.2e+003      \\       
    17     &   702    &     4638.53   &  9.16416e-005    &     -2e+004      \\       
    18     &   743    &     4569.17   &  2.25799e-005    &  -9.29e+003      \\       
    19     &   785    &     4502.31   &  7.32695e-005    &        -277      \\       
    20     &   827    &     4446.65   &   0.000227162    &       -86.5      \\       
    21     &   868    &     4199.75   &   0.000361625    &  -9.57e+003      \\       
    22     &   909    &     4140.98   &   6.2771e-005    &  -2.72e+003      \\       
    23     &   950    &     4052.51   &  8.45911e-005    &        -374      \\       
    24     &   991    &     3989.78   &   0.000138648    &       -44.9      \\       
    25     &  1032    &     3907.17   &   0.000104626    &         907      \\       
    26     &  1074    &     3856.12   &   0.000173797    &        -298      \\       
    27     &  1116    &     3779.25   &   0.000346633    &       -21.2      \\       
    28     &  1157    &     3645.96   &   0.000344504    &        -692      \\       
    29     &  1198    &      3619.6   &   0.000148309    &       -20.3      \\       
    30     &  1240    &     3539.96   &   0.000609224    &       -67.3      \\       
    31     &  1281    &     3204.79   &   0.000461207    &  -8.13e+003      \\       
    32     &  1322    &     3173.84   &   9.8575e-005    &  -2.55e+003      \\       
    33     &  1364    &     3039.27   &    0.00127777    &  -1.33e+003      \\       
    34     &  1406    &     2869.31   &     0.0301385    &       -28.1      \\       
    35     &  1447    &     2805.64   &    0.00227993    &  -3.43e+003      \\       
    36     &  1488    &     2252.13   &     0.0840027    &  -2.31e+003      \\       
    37     &  1529    &      1774.1   &      0.405937    &       -46.5      \\       
    38     &  1570    &     1438.04   &      0.390156    &      -0.399      \\       
    39     &  1612    &     1242.37   &      0.864944    &      -0.572      \\       
    40     &  1653    &     1166.62   &      0.702619    &       0.178      \\       
    41     &  1694    &     1124.36   &      0.492462    &      -0.158      \\       
    42     &  1735    &     1030.76   &       1.09544    &          -1      \\       
    43     &  1776    &     957.955   &      0.812872    &          -3      \\       
    44     &  1817    &     905.386   &      0.837528    &      -0.884      \\       
    45     &  1858    &     877.988   &        0.5546    &     -0.0476      \\       
    46     &  1899    &     856.077   &      0.793661    &      -0.164      \\       
    47     &  1940    &       847.3   &       1.14172    &    -0.00252      \\       
    48     &  1981    &     843.999   &      0.434118    &     -0.0237      \\       
    49     &  2022    &     842.438   &      0.780044    &     -0.0019      \\       
    50     &  2063    &     841.821   &      0.826689    &   4.11e-006      \\       
    51     &  2105    &     841.286   &       1.55349    &   -0.000208      \\       
    52     &  2146    &     840.828   &         1.651    &   -0.000259      \\       
    53     &  2187    &     840.559   &      0.904055    &   -0.000108      \\       
    54     &  2228    &     840.414   &      0.636094    &    0.000147      \\       
    55     &  2270    &     840.334   &       1.07033    &  -4.16e-005      \\       
    56     &  2311    &     840.311   &       1.03604    &  -6.22e-005      \\       
    57     &  2353    &      840.29   &       1.88839    &  -3.81e-005      \\       
    58     &  2394    &     840.274   &       1.85123    &  -1.93e-005      \\       
    59     &  2435    &     840.269   &       1.53297    &  -6.25e-006      \\       
    60     &  2476    &     840.266   &       1.20127    &  -9.11e-006      \\       
    61     &  2517    &     840.265   &       1.23825    &  -8.04e-006      \\       
    62     &  2558    &     840.265   &       1.02827    &   1.41e-005      \\       
    63     &  2599    &     840.265   &      0.736518    &  -1.31e-005      \\       
    64     &  2640    &     840.265   &      0.484504    &   1.86e-005      \\       
    65     &  2682    &     840.265   &      0.627666    &   -0.000102      \\       
    66     &  2720    &     840.265   &      0.313833    &   1.45e-005      \\       
    67     &  2758    &     840.265   &      0.107803    &  -7.09e-005      \\       
    68     &  2796    &     840.265   &    -0.0539014    &   1.81e-006      \\       
    69     &  2834    &     840.265   &      0.851352    &   7.68e-005      \\       
    70     &  2872    &     840.265   &      0.349962    &   4.96e-006      \\       
    71     &  2910    &     840.265   &      0.113811    &   2.09e-006      \\       
    72     &  2948    &     840.265   &     0.0382288    &   -0.000179      \\       
    73     &  2986    &     840.265   &     0.0191144    &   -0.000117      \\       
    74     &   3024   &     840.265   &     -0.0095572   &   -8.15e-006     \\\hline 
\end {tabular}}
\end{table}}
\clearpage

\section{Conclusions and Further Investigations}
In this report, we have documented a complete implementation of the flexible camera calibration method. Using 3 cameras, we have cross-validated that our camera calibration code is as good as the Microsoft code posted on the web page \cite{zhang98calibrationwebpage}. We have re-derived all the equations in \cite{zhang99calibrationinpaper,zhang99calibrationinreport} and corrected a technical error found in \cite{zhang99calibrationinpaper,zhang99calibrationinreport}. Compared to the work in \cite{zhang99calibrationinpaper,zhang99calibrationinreport}, where the feature location algorithms were not discussed, we have built a complete code for camera  calibration starting from the raw image acquisition step. A new method to effectively find the feature locations of the calibration object has been used in the code. More specifically, the scan line approximation algorithm is proposed to accurately determine the partitions of a given set of points. 

Based on our own platform and code, we can try some new ideas. In what follows, we will describe some of the efforts for the improved performance in camera calibration. 

\subsection{Error Reduction in Extraction of Feature Locations}
A big error source for calibration is the error in extraction of the feature locations and this is noticeable in Figure \ref{fig: extracted corners desktop}. There are a number of available feature extraction procedures developed \cite{chapter4,finalreportinStanford} and the procedure in Section \ref{section: extraction of feature locations} is our local line fitting method. Since the line fitting method is based on a small number of pixels, each noisy data point will strongly affect the fitting result. To improve the accuracy, an alternate method for line fitting is to use the {\it Hough Transform}, which is a standard tool in the domain of artificial vision for the recognition of straight lines, circles \cite{Song02sparsehough} and ellipses \cite{Song02sparsehough} and is particularly robust to missing and contaminated data \cite{Milan98imageprocessinganalysisandmachinevision,Song02sparsehough}. 

\begin{figure}[htb]
\centerline{\epsfxsize= 2.3in \epsffile{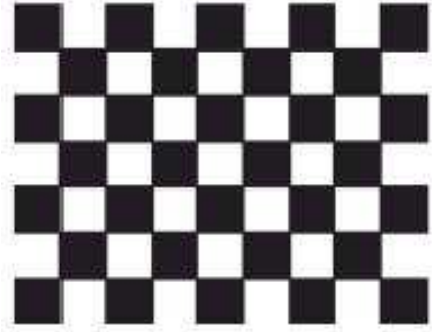}}
\caption {The checkerboard calibration object}
\label{fig: checkerboard calibration pattern}
\end{figure}

\begin{figure}[htb]
\centerline{\epsfxsize=4in \epsffile{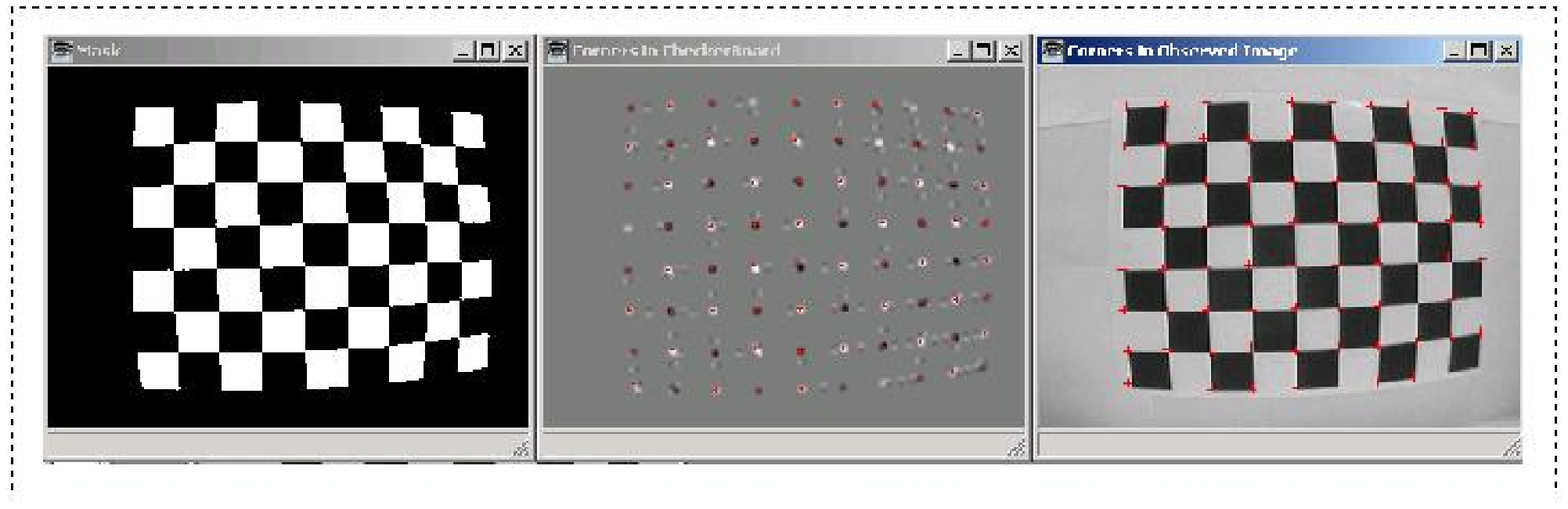}}
\caption {The result of convolving mask image with filter in Table \ref{table: convolving window for checkerboard calibration pattern} (1)}
\label{fig: checkerboard filter output correct}
\end{figure}

\begin{figure}[htb]
\centerline{\epsfxsize=4in \epsffile{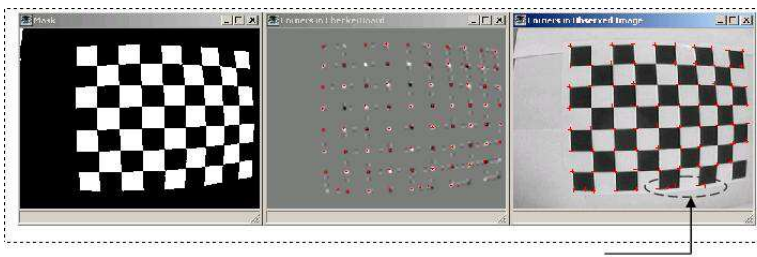}}
\caption {The result of convolving mask image with filter in Table \ref{table: convolving window for checkerboard calibration pattern} (2)}
\label{fig: checkerboard filter output wrong}
\end{figure}

Using the calibration object in Figure \ref{fig: calibration object}, one shortcoming is that when choosing different threshold values to convert the observed intensity image to the binary image, the threshold may cause the squares to shrink or enlarge, which will in turn affect the feature localization. In \cite{intel,chapter4}, the checkerboard pattern is proposed as the calibration object. The advantage of the checkerboard pattern, as shown in Figure \ref{fig: checkerboard calibration pattern}, is that its corners are localizable independent of the linearity of the image response. That is, applying a nonlinear monotonic function to the intensity values of the checkerboard image, such as gamma correction, does not affect the corner localizations. Using the checkerboard calibration pattern, the feature locations can be extracted by first convolving the mask image (obtained by thresholding the input PGM format image) with the window shown in Table \ref{table: convolving window for checkerboard calibration pattern} \cite{chapter4}. Since this filter itself resembles a checkerboard pattern, it gives a strong response, positive or negative, depending on which type of corner when centered over a checkerboard corner. The filter output of the mask image produces an image where the checkerboard corners appear as white or black dots (See Figure \ref{fig: checkerboard filter output correct} and \ref{fig: checkerboard filter output wrong}). In \cite{chapter4}, localizing a particular checkerboard corner after the filter convolution is said to be easily accomplished by locating the point of maximum positive or negative filter response. Though not so straightforward, the checkerboard corners can be extracted by adding two extra steps. In the image of filter output, the values of most pixels are 0 except in the small region around corners or edges, where pixels can be mostly positive or negative depending on what kind of corners. So, the first step added is to get the region map and in each region locate the pixel whose filter response is the strongest. Doing this way, the output pixels may include some noisy pixels around the edges and the second step is to pick up a certain number of pixels with the strongest filter response from all regions based on the known knowledge of how many corners in the given pattern. The filter output and extracted corners are shown in Figures \ref{fig: checkerboard filter output correct} and \ref{fig: checkerboard filter output wrong}. As can be seen from Figure \ref{fig: checkerboard filter output wrong}, the current simple method is not robust enough and the feature locations can not always be extracted accurately. Some post-processing algorithms are thus needed. Other algorithms exist such as 1) using Canny edge detector to find the edge pixels, 2) globally fit curves to the edge pixels and 3) find their intersections \cite{finalreportinStanford} (also using the checkerboard pattern). However, for both calibration objects, in order to achieve a sub-pixel accuracy, a series of post-processing algorithms are necessary. 

\begin{table}[htb]
\centering
\caption{Convolution Window for the Checkerboard Calibration Object}
\label{table: convolving window for checkerboard calibration pattern}
\bigskip
{\begin {tabular}{|c|c|c|c|c|c|c|}\hline
-1 & -1 & -1 & 0 &  1 &  1 &  1\\ \hline
-1 & -1 & -1 & 0 &  1 &  1 &  1\\ \hline
-1 & -1 & -1 & 0 &  1 &  1 &  1\\ \hline
 0 &  0 &  0 & 0 &  0 &  0 &  0\\ \hline
 1 &  1 &  1 & 0 & -1 & -1 & -1\\ \hline
 1 &  1 &  1 & 0 & -1 & -1 & -1\\ \hline
 1 &  1 &  1 & 0 & -1 & -1 & -1
\\\hline
\end {tabular}}
\end{table}

\clearpage
\subsection{Speed Up the Nonlinear Optimization}
Currently, we put totally $7+6N$ parameters into the nonlinear optimization routine to get the final estimations. In Matlab, this is absolutely time consuming. Can we make nonlinear optimization faster? 

\begin{figure}[htb]
\begin{minipage}{\textwidth}
\setcounter{mpfootnote}{\value{footnote}}
\renewcommand{\thempfootnote}{\arabic{mpfootnote}}
\centerline{\epsfxsize=5.5in \epsffile{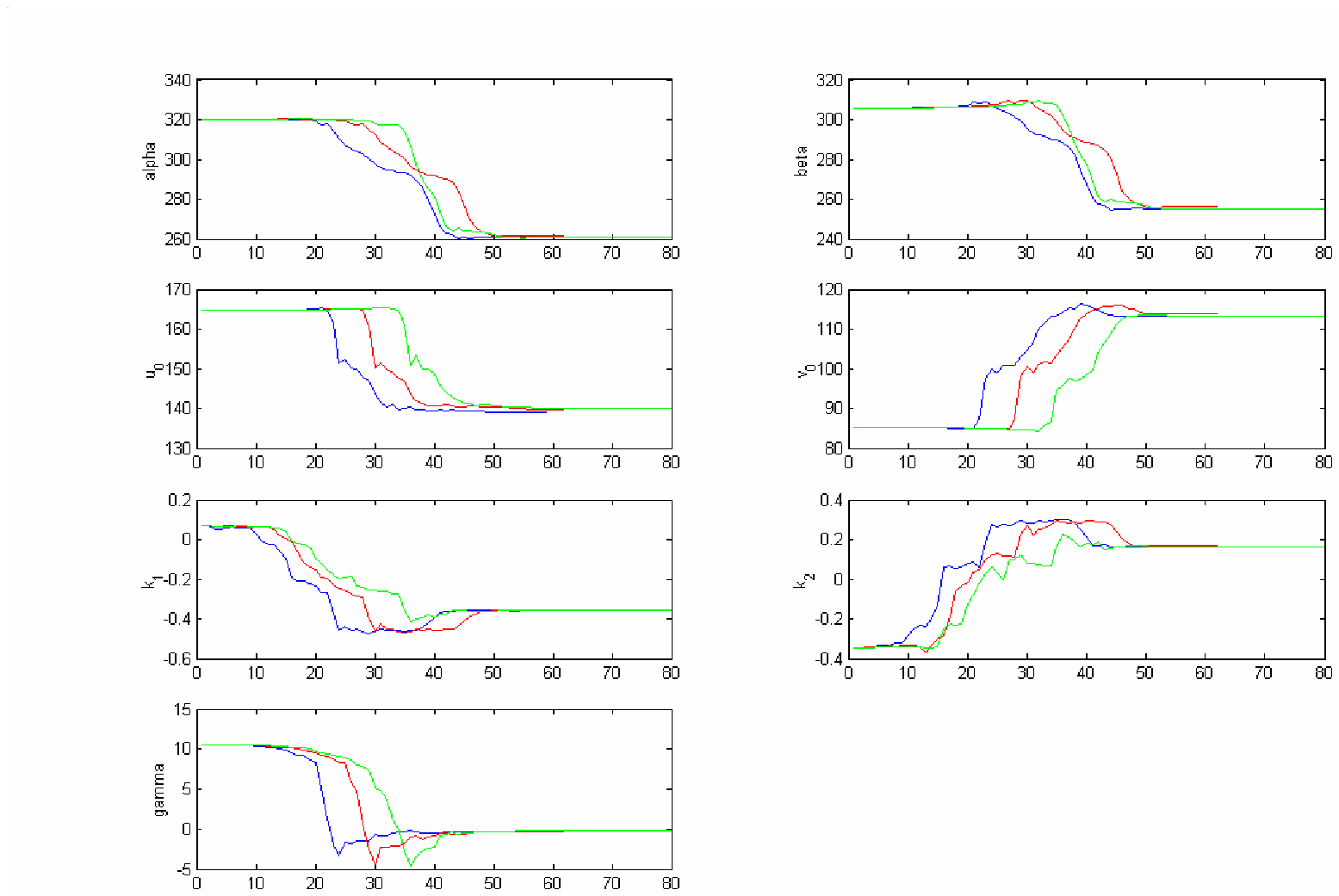}}
\caption [Intrinsic parameters through nonlinear optimization iterations] {Intrinsic parameters through nonlinear optimization iterations \footnote{Blue color is for 3 images, red for 4 images, and green for 5 images}}
\label{fig: optimization 345}
\end{minipage}
\end{figure}

Apply 5 images to get the initial guess of camera parameters. Then use 3, 4, and 5 image data for nonlinear optimization respectively. The 7 intrinsic parameters at each iteration are plotted in Figure \ref{fig: optimization 345}, where blue color is for the nonlinear optimization using 3 images, similarly red for 4 images, and green for 5 images. It is clear that the final estimation are very consistent with each other and using only 3 images for nonlinear optimization is absolutely time saving.

\subsubsection{Incremental Nonlinear Optimization}
To speed up the nonlinear optimization, one idea we came out with is to apply the first 3 images to nonlinear optimization routine, and then use the output 7 intrinsic parameters (treat these 7 parameters as unchanged) to optimize the extrinsic parameters for the $4^{th}$ and $5^{th}$ image. At last, put all the estimated parameters again into the nonlinear optimization routine to get the final estimates. Doing this way, we call it incremental nonlinear optimization. Disappointedly, compared with the one-step optimization method, our initial attempt shows that the incremental method can only reach the same performance in terms of the accuracy and the consuming time. 

\subsubsection{Constrained Nonlinear Optimization}
Another idea that might speed up the nonlinear optimization is to add some constrains, such as:
\begin{itemize}
\item{$u_0$ is close to half of image's length}
\item{$v_0$ is close to half of image's width}
\item{angle between two image axes is close to 90 degree}
\item{$\alpha, \ \beta$ are greater than 0}
\end{itemize}
We shall investigate this idea later.

\subsection{Radial Undistortion}
According to the radial distortion model in Equation (\ref{eqn: radial distortion}), radial distortion can be understood to perform in one of the  following two ways: 
\begin{itemize}
\item{Perform distortion in the camera frame, then transform to the image plane} 
\begin{eqnarray}
{\left [\matrix {
x \cr
y}\right]} \rightarrow
{\left [\matrix {
x' \cr
y'}\right]} \rightarrow
{\left [\matrix {
u_d \cr
v_d}\right]} \nonumber
\end{eqnarray}
\item{Transform to the image plane, then perform distortion in the image plane} 
\begin{eqnarray}
{\left [\matrix {
x \cr
y}\right]} \rightarrow
{\left [\matrix {
u \cr
v}\right]} \rightarrow
{\left [\matrix {
u_d \cr
v_d}\right]}\nonumber
\end{eqnarray}
\end{itemize}
To explain, let $f(r) = 1 + k_1 r^2 + k_2 r^4$, $r^2 = x^2 + y^2$, we know that
\begin{eqnarray}
u_d &=& (u-u_0) \ f(r) + u_0 \nonumber \\
    &=& \alpha \ x f(r) + \gamma \ y f(r) + u_0 \nonumber\\
    &=& \alpha \ x' + \gamma \ y' + u_0, \\
v_d &=& (v-v_0) \ f(r) + v_0 \nonumber\\
    &=& \beta \ y'  + v_0, 
\end{eqnarray}
where
\begin{eqnarray}
\label{eqn: x'y' and xy}
\begin{array}{l}
x' = x \ f(r) \\
y' = y \ f(r).
\end{array} 
\end {eqnarray}

{\it Radial undistortion} is to extract $(u,v)$ from $(u_d, v_d)$, which can be done by extracting $(x,y)$ from $(x', y')$. The following derivation shows the problem when trying to extract $(x,y)$ from $(x', y')$ using two distortion coefficients $k_1$ and $k_2$. From $(u_d, v_d)$, we can calculate $(x', y')$ by 
\begin{eqnarray}
{\left [\matrix {
x' \cr
y' \cr
1}\right]} = A^{-1} {\left [\matrix {
u_d \cr
v_d \cr
1}\right]}
= {\left [\matrix {
\frac{1}{\alpha} & -\frac{\gamma}{\alpha \beta} & -\frac{u_0}{\alpha}+\frac{v_0 \gamma}{\alpha \beta} \cr
0 & \frac{1}{\beta} & -\frac{v_0}{\beta} \cr
0 & 0 & 1 }\right]} {\left [\matrix {
u_d \cr
v_d \cr
1}\right].}
\end{eqnarray}
Now the problem becomes to extract $(x, y)$ from $(x', y')$. According to Equation (\ref{eqn: x'y' and xy}),
\begin{eqnarray}
\begin{array}{l}
x' = x \ f(r) = x \ [1+k_1(x^2+y^2) + k_2(x^2+y^2)^2]  \\
y' = y \ f(r) = y \ [1+k_1(x^2+y^2) + k_2(x^2+y^2)^2]. 
\end{array}
\end{eqnarray}
Let $c = y'/x' = y /x$, we have $y = cx$ where $c$ is a constant. Substituting $y = cx$ into the above equation, gives
\begin{eqnarray}
\label{eqn: undistortion case(1)}
x' &=& x \ [1+k_1(x^2+c^2x^2)+k_2(x^2+c^2x^2)^2] \nonumber\\
   &=& x + k_1 (1+c^2)x^3+k_2(1+c^2)^2x^5 .
\end{eqnarray}
Let $f_1(x) = x + k_1 (1+c^2)x^3+k_2(1+c^2)^2x^5$, then $f_1(-x) = - f_1(x)$ so $f_1(x)$ is an odd function. The analytical solution of Equation (\ref{eqn: undistortion case(1)}) is not a trivial task. It is said that this problem is still open (of course, we can use numerical method to solve it). But if we set $k_2 = 0$, the analytical solution is available and radial undistortion can be done easily. In \cite{Undistortionchapter}, for the same practical purpose, only one distortion coefficient $k_1$ is used to approximate radial distortion. 

Here, let's consider the distortion model again. There are two questions we are of particular interests. The first question is: can we choose
\begin{equation}
\hspace{-1.49cm} {\tt radial \ distortion \ model \ case _2:} \ f(r) = 1 + k_1 r^2
\end{equation}
instead of 
\begin{equation}
{\tt radial \ distortion \ model \ case_1:} \ f(r) = 1 + k_1 r^2 + k_2 r^4 \ ?
\end{equation}
What can be the performance degradation? The second question is: can we model radial distortion in other way so that we can achieve good and reasonable accuracy along with easy analytical undistortion? To answer the first question, we can re-optimize the parameters with only one distortion coefficient $k_1$. Recall that the initial guess for radial distortion is done after having estimated all other parameters and just before nonlinear optimization. So we can reuse the estimated parameters and choose the initial guess for $k_1$ to be 0. For the second question, let's take 
\begin{equation}
\hspace{-0.5cm} {\tt radial \ distortion \ model \ case_3:} \ f(r) = 1 + k_1 r + k_2 r^2, 
\end{equation} 
which is a function only related to radius $r$.  The motivation of choosing this radial distortion model is that the resultant approximation of $x'$ is also an odd function of $x$, as can be seen next.

When $F(r) = r f(r) = r (1 + k_1 r + k_2 r^2)$, we have
\begin{eqnarray}
\begin{array}{l}
x' = x \ f(r) = x \ [(1 + k_1 r + k_2 r^2)]  \\
y' = y \ f(r) = y \ [(1 + k_1 r + k_2 r^2)]. 
\end{array}
\end{eqnarray}
Again let $c = y'/x' = y /x$, we have $y = cx$ where $c$ is a constant. Substituting $y = cx$ into the above equation, gives
\begin{eqnarray}
\label{eqn: distortion model 3}
x' &=& x \ \left[ 1+k_1 \sqrt{x^2 + c^2x^2} + k_2(x^2 + c^2x^2)\right] \nonumber\\
   &=& x \ \left[ 1+k_1 \sqrt{1 + c^2} \ {\tt sign}(x) x + k_2(1 + c^2)x^2 \right] \nonumber\\
   &=& x + k_1 \sqrt{1 + c^2} \ {\tt sign}(x) \ x^2 + k_2(1 + c^2) \ x^3.
\end{eqnarray}
Let $f_3(x) = x + k_1 \sqrt{1 + c^2} \ {\tt sign}(x) \ x^2 + k_2(1 + c^2) \ x^3$, $f_3(x)$ is also an odd function. 

Now, we want to compare the three radial distortion models based on the final value of objective function after nonlinear optimization by Matlab function {\tt fminunc}. Using Microsoft images, desktop images, and ODIS images, the objective function, 5 intrinsic parameters, and distortion coefficients are shown in Table \ref{table: Comparison of Three Distortion Models}. The results in Table \ref{table: Comparison of Three Distortion Models} show that the objective function of Model$_3$ is always greater than that of Model$_1$, but smaller than that of Model$_2$. This is consistent with our anticipation. 

The benefits of using radial distortion model$_3$ are:
\begin{enumerate}
\item[(1)]{Low order fitting, better for fixed-point implementation}
\item[(2)]{Explicit inverse function with no numerical iterations}
\item[(3)]{Better accuracy than radial distortion model$_2$}
\end{enumerate}

{\small
\begin{table}[htb]                                                                 
\centering                                                                         
\caption{Comparison of Three Distortion Models}
\label{table: Comparison of Three Distortion Models}
\renewcommand{\arraystretch}{1}
\bigskip
{\begin {tabular}{|c||r|r|r||r|r|r||r|r|r|}\hline
& \multicolumn{3}{|c||}{\bf Microsoft Images} & \multicolumn{3}{|c||}{\bf Desktop Images} & \multicolumn{3}{|c|}{\bf ODIS Images} \\\hline
{\bf Model} & ${\tt \#1}$ & $\#2$ & $\#3$ & $\#1$ & $\#2$ & $\#3$ & $\#1$ & $\#2$ & $\#3$ \\\hline
{\bf $J$} &  {\bf 144.88}   & {\bf 148.279}  & {\bf 145.659}   & {\bf 778.9768}   &  {\bf 904.68}  & {\bf 803.307}   & {\bf 840.2650}  &  {\bf 933.098} & {\bf 851.262}  \\\hline
$\alpha$ &  832.5010 & 830.7340 & 833.6623  & 277.1457   & 275.5959 & 282.5664  &260.7636   & 258.3206    & 266.0861 \\
$\gamma$ &    0.2046 &   0.2167 & 0.2074    &    -0.5730 &  -0.6665 &   -0.6201 &   -0.2739 &    -0.5166  &  -0.3677 \\
$u_0$    &  303.9584 & 303.9583 & 303.9771  &   153.9923 & 158.2014 &  154.4891 &  140.0564 &   137.2155  & 139.9177 \\
$\beta$  &  832.5309 & 830.7898 & 833.6982  &   270.5592 & 269.2307 &  275.9040 &  255.1465 &   252.6869  & 260.3145 \\
$v_0$    &  206.5879 & 206.5692 & 206.5520  &   119.8090 & 121.5254 &  120.0952 &  113.1723 &   115.9295  & 113.2417 \\
$k_1$    &   -0.2286 &  -0.1984 &  -0.0215  &    -0.3435 &  -0.2765 &   -0.1067 &   -0.3554 &    -0.2752  &  -0.1192 \\
$k_2$    &    0.1903 &        0 &  -0.1565  &     0.1232 &        0 &   -0.1577 &    0.1633 &          0  &  -0.1365 \\\hline
\end {tabular}}                                                                    
\end{table}}

To extract $x$ from $x'$ in Equation (\ref{eqn: distortion model 3}), let's first assume $x>0$. Then there is an unique solution $\in \mathcal{R}$, denoted by $x_{+}$, that satisfies Equation (\ref{eqn: distortion model 3}) but may or may not coincide with the assumption. If $x_{+} > 0$, $x=x_{+}$, otherwise, $x = x_{-}$, which is the solution ($\in \mathcal{R}$) for the case when $x<0$. 


\subsection{Possible Applications of Camera Calibration in ODIS Missions}

\subsubsection{Visual Servoing Continuity}
The wireless visual servoing in \cite{berkemeier01visualservo,lili02visualservo} is performed by an uncalibrated camera. The experiment results do not match the theoretical simulation results satisfactorily because the parameters of the camera model are roughly set. With the camera calibration work presented here, we should be able to identify the parameters needed in simulation. 

\subsubsection{Estimation of the Pan/Tilt Angles of the ODIS Camera}
The following steps can be applied to estimate the pan/tilt angles of the ODIS camera with respect to the ODIS body fixed coordinate system:
\begin{itemize}
\item{put the calibration object perpendicular to the ground, and let the bottom-left corner be $[0,0,0]$ of the world coordinate system}
\item{put ODIS in front of the calibration object, and be sure to make its $x_B$ and $y_B$ axes parallel to those of world coordinate system}
\item{take a picture of calibration object using ODIS camera}
\item{estimate homography $H$}
\item{estimate 6 extrinsic parameters using intrinsic parameters we already know and the homography estimated in the last step}
\item{from the 3 angular vector, we can estimate the pan/tilt angles of ODIS camera}
\end{itemize}

\subsubsection{Non-iterative Yellow Line Alignment with a Calibrated Camera}
Instead of our previous yellow line alignment method described in \cite{berkemeier01visualservo,lili02visualservo}, can we align to yellow line with a non-iterative way using a calibrated camera? The answer is yes. Let's begin with a case when only ODIS's yaw and $x, y$ positions are unknown while ODIS camera's pan/tilt angles are unchanged since calibration. The problem is described in detail in Table \ref{table: visual servo task1}. 

\begin{table}[htb]
\centering
\caption{Task of Yellow Line Allignment Using Calibrated ODIS Camera}
\label{table: visual servo task1}
\bigskip
{\begin {tabular}{|l|}\hline
{\bf Given}:    \ 3D locations of yellow line's two ending points\\
\indentintabular \indentintabular ODIS camera's pan/tilt angles\\
\indentintabular \indentintabular ODIS camera's intrinsic parameters\\
\indentintabular \indentintabular ODIS camera's radial distortion model and distortion coefficeints \\
\indentintabular \indentintabular The two projected points in image plane using ODIS camera \\
{\bf Find}: \ \ ODIS's actural yaw and $x, y$ positions \\\hline
\end {tabular}}
\end{table}

\begin{figure}[htb]
\centerline{\epsfxsize=4in \epsffile{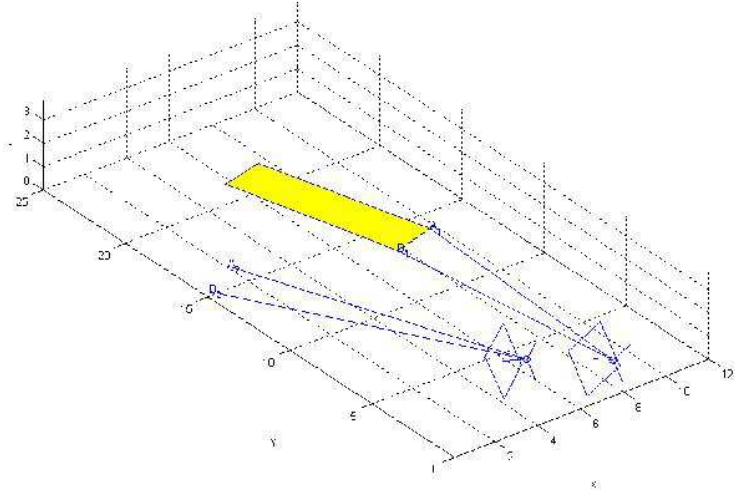}}
\caption {The task of yellow line alignment}
\label{fig: visual servo1}
\end{figure}

Knowing that a change in ODIS's yaw angle only results in a change of angle $c$ in the $ZYZ$ Euler angles $(a,b,c)$. So, when using $ZYZ$ Euler angles to identify ODIS camera's orientation, we will assume the first two varialbes $a, b$ are unchanged. In Figure \ref{fig: visual servo1}, after some time of navigation, the robot thinks it is at Position 1. Then it sees the yellow line, whose locations in 3D world reference frame are known from map (denoted by $P_A^w$ and $P_B^w$). After extracted the correspoinding points in image plane of the yellow line's two ending points, we can calculate the undistorted image points and thus recover the 3D locations of the two ending points (denoted by $P_{AA}^w$ and $P_{BB}^w$), using ODIS camera's 5 intrinsic parameters and 2 radial distortion coefficients. From the difference between the yellow line's actual locations in map and the recovered locations, the deviation in the robot's $x, y$ positions and yaw angle can be calculated. 

Let $(x_1, y_1)$ and $(x_2, y_2)$ be the corresponding points in camera frame of yellow line's two ending points. Let $R_2$ and ${\bf t}_2$ be the rotation matrix and translation vector at position 1 (where the vehicle thinks it is at), similarly $R_1$ and ${\bf t_1}$ at position 2 (the true position and orientation), we can write $R_2 = \Delta R \cdot R_1$ and ${\bf t}_2 = {\bf t}_1 + \Delta {\bf t}$, where $\Delta R$ and $\Delta {\bf t}$ are the deviation in orientation and translation. If the transform from the world reference frame to the camera frame is $P^c = R^{-1} (P^w - {\bf t})$, first we will calculate $P_{AA}^w$ and $P_{BB}^w$. Let $P_{AA}^w = [X_{AA}^w, Y_{AA}^w, 0]$, we have
\begin{eqnarray}
P^c = \left [\matrix{X^c \cr Y^c \cr Z^c }\right] = R_2^{-1} \left[\matrix{X_{AA}^w - {\bf t}_{21}\cr Y_{AA}^w - {\bf t}_{22}\cr -{\bf t}_{23}}\right].
\end{eqnarray}
Since \begin{eqnarray}
\frac{X^c}{x_1} = \frac{Y^c}{y_1} = \frac{Z^c}{1},
\end{eqnarray}
we have two equations containing two variables and $P_{AA}^w$ can be calculated out. In a same way, we can get $P_{BB}^w$. 

Knowing $P_{AA}^w$ and $P_{BB}^w$, we have
\begin{eqnarray}
\label{eqn: point A}
\lambda \left [ \matrix {
x_1 \cr
y_1 \cr
1
} \right] = R_2^{-1} \Delta R (P_A^w - {\bf t}_1) = R_2^{-1} (P_{AA}^w - {\bf t}_2), 
\end{eqnarray}
where $\lambda$ is a scaling factor. 
From Equation (\ref{eqn: point A}), we get 
\begin{eqnarray}
\label{eqn: A - AA}
R_2^{-1} [\Delta R (P_A^w - {\bf t}_1) - P_{AA}^w + {\bf t}_2] = 0.
\end{eqnarray}
Similarly, we get
\begin{eqnarray}
\label{eqn: B - BB}
R_2^{-1} [\Delta R (P_B^w - {\bf t}_1) - P_{BB}^w + {\bf t}_2] = 0. 
\end{eqnarray}
Using the above two equations, we have $(P_{AA}^w - P_{BB}^w) = \Delta R \ (P_A^w - P_B^w)$, where $\Delta R$ is of the form 
\begin{eqnarray}
\Delta R = \left [ \matrix {
\cos(\Delta \theta) & -\sin(\Delta \theta) & 0 \cr
\sin(\Delta \theta) &  \cos(\Delta \theta) & 0 \cr
0 & 0 & 1
} \right].
\end{eqnarray}
So, $\Delta \theta$ is just the rotation angle from vector $P_A^w \rightarrow P_B^w$ to vector $P_{AA}^w \rightarrow P_{BB}^w$. Knowing $\Delta R$, ${\bf t}_1$ becomes ${\bf t}_1 = P_A^w - {\Delta R}^{-1} P_{AA}^w + {\Delta R}^{-1} R_2 {\bf t}_2$. \\

{\bf Remaining question}: 
\begin{itemize}
\item{Choose $x$ from 6 possible values in radial undistortion for distortion Model$_3$}
\item{If ODIS camera's pan/tilt angles are changed due to user commanded tasks and are not available, what can we do for yellow line alignment?}
\end{itemize}

\subsection{Tidy Up of Our Camera Calibration Code}

\subsubsection{Singularity Reminder for Robustness in Camera Calibration}
When moving the camera or the calibration object to capture images, if the relative orientation between the calibration object and the camera does not change, singularity will occur. We will add one function in the code that can pop up information telling the user which two images may cause the singularity problem. We will also design an automatic procedure to discard the image that cause the singularity problem and use the remaining images for calibration, if the number of remaining images is greater than 2. To do so, a numerical indicator will be used to predict the degree of singularity. 

\subsubsection{Two Complete Sets of Codes: Matlab and C/C++}
Currently, all of the codes are in C/C++ except the nonlinear optimization routine. We should be able to find some open source code in C/C++ for nonlinear optimization so that we can have a complete set of C/C++ code for camera calibration. Meanwhile, we would like to have a pure Matlab version by CMEX facility provided in Matlab. 

\appendix{}
\section{Nonlinear Optimization using TOMLAB}
Using the nonlinear optimization problem defined in section \ref{section: nonlinear optimization} and with the data set for calibration of ODIS camera, we can evaluate the performance of TOMLAB NLPLIB Toolbox with the results reported in the literature, such as Matlab Optimization Toolbox. TOMLAB NLPLIB toolbox is a set of Matlab solvers, test problems, graphical and computational utilities for unconstrained and constrained optimization, quadratic programming, unconstrained and constrained nonlinear least squares, box-bounded global optimization, global mixed-integer nonlinear programming, and exponential sum model fitting \cite{TOMLABmanual1,TOMLABmanual2}. In TOMLAB, the routine {\tt ucSolve} implements a prototype algorithm for unconstrained or constrained optimization with simple bounds on the variables, i.e. solves the problem
\begin{eqnarray}
\begin{array}{l}
\min_x \ f(x) \\
\ s.t. \ \ x_L \le x  \le x_U,
\end{array}
\end{eqnarray}
where $x, x_L,$ and $x_U \in {\mathcal R}^n$ and $f(x) \in \mathcal{R}$.
The algorithms implemented in {\tt ucSolve} are listed in Table \ref{table: ucsolve algorithms}. 
\begin{table}[h]
\centering
\caption{Algorithms Implemented in {\tt ucSolve}}
\label{table: ucsolve algorithms}
\bigskip
{\begin {tabular}{|c|l|}\hline
{\bf Variable} & {\bf Algorithm} \\[1ex]\hline 
0 & 	Default algorithm (BFGS or Newton)\\\hline
1 & 	Newton with subspace minimization, using {\tt SVD}\\\hline
2 & 	Safeguarded BFGS with standard inverse Hessian update\\\hline
3 & 	Safeguarded BFGS with Hessian update, and SVD or LU to solve\\\hline
4 & 	Safeguarded DFP  with standard inverse Hessian update\\\hline
5 & 	Safeguarded DFP  with Hessian update, and SVD or LU to solve \\\hline
6 & 	Fletcher-Reeves CG \\\hline
7 & 	Polak-Ribiere CG \\\hline
8 & 	Fletcher conjugate descent CG-method \\\hline
\end {tabular}}
\end{table}

{\bf Performance comparison of these two nonlinear optimization toolboxes are based on the number of iterations and consuming time to reach the same or similar value for objective function. It should also be kept in mind that the comparison should go to the case when both these two toolboxes implement the same or similar algorithms}. 

In Matlab, we choose function {\tt fminunc} that uses BFGS Quasi-Newton method with a mixed quadratic and cubic line search procedure (because we do not provide gradient information and we thus use Medium-Scale Optimization algorithm) \cite{MATLABoptimizationtoolbox}. Correspondingly, in TOMLAB, we pick up algorithm $\# 0$ with Quadratic Interpolation or Cubic Interpolation LineSearchType. This is because {\tt ucSolve} does not have a mixed quadratic and cubic line search procedure and the above two methods are what available. The command lines are:
\begin{itemize}
\item{Matlab {\tt fminunc} command line:}\\
{\small
options = optimset(`Display',`iter', `LargeScale', `off', `TolX', $10^{-5}$, `TolFun', $10^{-5}$); \\
x = fminunc(@myfun, x0, options);}\\
{\footnotesize Note: when `LineSearchType'' is not specified, the default value is ``quadcubic''.} 

\item{TOMLAB {\tt ucSolve} command line:}\\
{\small
Prob = conAssign(`myfun', [], [], [], [], [], `calibration', x0);\\
Prob.Solver.Alg = 0;\\ 
Prob.PriLevOpt = 2;\\
Result = ucSolve(Prob);}\\
{\footnotesize Note: to specify the line search method for {\tt ucSolve$_0$}, parameter ``LineAlg'' is changed inside routine ``LineSearch.m''.}
\end{itemize}

Table \ref{table: Matlab fminunc} and \ref{table: TOMLAB ucsolve0} show the optimization results for Matlab {\tt fminunc} and TOMLAB {\tt ucSolve} respectively, where each column represents Iteration Number, Objective Function at Each Iteration, Time at Each Iteration (obtained by Matlab command ``rem(now,1)''), and Parameters Estimated respectively (in both tables, the number with a left arrow on the side is the total running time till current iteration). From these two tables, we observe: 

\begin{itemize}
\item{Both functions converge}
\item{Both functions converge to close values for both objective function and estimated parameters}
\item{Matlab converge with less iterations and less time, slightly having advantage over TOMLAB}\\\\
For line search method, it is a common sense that quadratic interpolation that involves a data fit to the form $ax^2+bx+c$ should take less time than cubic interpolation $ax^3+bx^2+cx+d$. In a same manner, a mixed quadcubic interpolation might lie somewhere between quadratic and cubic interpolations. Thinking this way, Matlab function demonstrates faster convergent speed over TOMLAB.
\end{itemize}

{\footnotesize
\begin{table}[htb]
\centering
\caption{Matlab {\tt fminunc} with {\tt LineSearchType = QuadCubic}}
\label{table: Matlab fminunc}
\renewcommand{\arraystretch}{0.9}
\bigskip
	{\begin {tabular}{|r|l|r|r|}\hline
	\multicolumn{4}{|c|}{{\bf Matlab {\tt fminunc}}}\\\hline
	{\bf $\#$} & {\bf $\ \ \ \ f(x)$} & {\bf Time ($10^{-1}$)} & {\bf Params} \\\hline
	 1  &   5872.79   &  5.309561  &    260.7636  \\
	 2  &   5849.4    &  5.310881  &     -0.2739  \\
	 3  &   5841.17   &  5.312219  &    140.0564  \\
	 4  &   5734.46   &  5.313533  &    255.1465  \\
	 5  &   5703.36   &  5.314850  &    113.1723  \\
	 6  &   5524.51   &  5.316187  &     -0.3554  \\
	 7  &   5505.25   &  5.317472  &      0.1633  \\
	 8  &   5286.49   &  5.318820  &     -1.8822  \\
	 9  &   5252.79   &  5.320125  &      2.8795  \\
	10  &   5147.55   &  5.321412  &     -1.9377  \\
	11  &   5123.77   &  5.322762  &      9.4005  \\
	12  &   5111.44   &  5.324098  &     -9.6023  \\
	13  &   5085.05   &  5.325382  &    -24.9324  \\
	14  &   5033.1    &  5.326730  &     -1.5881  \\
	15  &   5017.1    &  5.328034  &      2.8809  \\
	16  &   4875.16   &  5.329351  &     -1.6277   \\
	17  &   4638.53   &  5.330698  &      7.3082  \\
	18  &   4569.17   &  5.332003  &    -12.8265  \\
	19  &   4502.31   &  5.333320  &    -25.2393  \\
	20  &   4446.65   &  5.334666  &     -1.3550  \\
	21  &   4199.75   &  5.335973  &      2.8024  \\
	22  &   4140.98   &  5.337258  &     -1.3853  \\
	23  &   4052.51   &  5.338576  &     10.6469  \\
	24  &   3989.78   &  5.339881  &     -9.5356  \\
	25  &   3907.17   &  5.341169  &    -17.4115  \\
	26  &   3856.12   &  5.342516  &     -1.3737  \\
	27  &   3779.25   &  5.343853  &      2.8125  \\
	28  &   3645.96   &  5.345139  &     -1.3938  \\
	29  &   3619.6    &  5.346456  &     10.2720  \\
	30  &   3539.96   &  5.347795  &    -10.2214  \\
	31  &   3204.79   &  5.349080  &    -19.7591  \\
	32  &   3173.84   &  5.350395  &     -0.8848  \\
	33  &   3039.27   &  5.351733  &      2.6428  \\
	34  &   2869.31   &  5.353052  &     -0.8422  \\
	35  &   2805.64   &  5.354367  &      8.8168  \\
	36  &   2252.13   &  5.355673  &     -9.1196  \\
	37  &   1774.1    &  5.356961  &    -17.3228  \\
	38  &   1438.04   &  5.358281  &              \\
	39  &   1242.37   &  5.359631  &              \\
	40  &   1166.62   &  5.360914  &              \\
	41  &   1124.36   &  5.362229  &              \\
	42  &   1030.76   &  5.363537  &              \\
	43  &   957.955   &  5.364837  &              \\
	44  &   905.386   &  5.366189  &              \\
	45  &   877.988   &  5.367488  &              \\
	46  &   856.077   &  5.368767  &              \\
	47  &   847.3     &  5.370106  &              \\
	48  &   843.999   &  5.371396  &              \\
	49  &   842.438   &  5.372726  &              \\
	50  &   841.821   &  5.374871  &              \\
	51  &   841.286   &  5.376304  &              \\
	52  &   840.828   &  5.377721  &              \\
	53  &   840.559   &  5.379342  &              \\
	54  &   840.414   &  5.380722  & $\leftarrow${\bf 0.0071161}\\
	55  &   840.334   &  5.382321  &              \\
	56  &   840.311   &  5.383757  &              \\
	57  &   840.29    &  5.385097  &              \\
	58  &   840.274   &  5.386385  &              \\
	59  &   840.269   &  5.387730  &              \\
	60  &   840.266   &  5.389072  &              \\
	61  &   840.265   &  5.390453  &              \\
	62  &{\bf 840.265}&  5.391947  & $\leftarrow${\bf 0.0082386}\\\hline
	\end {tabular}}
\end{table}}

{\footnotesize
\begin{table}[htb]
\centering
\caption{TOMLAB {\tt ucSolve} Algorithm$_0$}
\label{table: TOMLAB ucsolve0}
\renewcommand{\arraystretch}{0.9}
\bigskip
{\begin {tabular}{|r|l|r|r|l|r|r|}\hline
\multicolumn{7}{|c|}{{\bf TOMLAB {\tt ucSolve}$_0$}}\\\hline
\multicolumn{4}{|c|}{{\tt LineSearchType = Quadratic Interpolation}}& \multicolumn{3}{c|}{{\tt LineSearchType = Cubic Interpolation}}\\\hline
{\bf $\#$}&{\bf $\ \ \ \ f(x)$}&{\bf Time ($10^{-1}$)}&{\bf Params}&{\bf $\ \ \ \ f(x)$}&{\bf Time ($10^{-1}$)} & {\bf Params} \\\hline
 0 &   5872.79  &   7.905221  & 260.5232   &   5872.79   &  5.607743  &  260.6613 \\
 1 &   5850.93  &   7.906629  &  -0.2585   &   5850.93   &  5.614631  &   -0.2650 \\
 2 &    5842.7  &   7.908060  & 139.4244   &   5842.81   &  5.623759  &  139.4470 \\
 3 &   5736.13  &   7.909437  & 254.8903   &   5750.96   &  5.630199  &  255.0380 \\
 4 &   5705.03  &   7.910831  & 113.0924   &   5728.82   &  5.637519  &  113.0792 \\
 5 &   5589.15  &   7.912188  &  -0.3549   &   5616.05   &  5.642693  &   -0.3553 \\
 6 &   5573.78  &   7.913598  &   0.1630   &   5598.15   &  5.650420  &    0.1635 \\
 7 &   5370.21  &   7.914950  &  -1.8919   &   5406.89   &  5.656750  &   -1.8910 \\
 8 &   5306.06  &   7.916366  &   2.8786   &   5365.39   &  5.663024  &    2.8785 \\
 9 &   5231.59  &   7.917751  &  -1.9471   &   5287.34   &  5.669369  &   -1.9462 \\
10 &   5208.66  &   7.919179  &   9.3391   &   5271.08   &  5.678234  &    9.3412 \\
11 &   5195.82  &   7.920595  &  -9.6113   &   5255.9    &  5.685982  &   -9.6124 \\
12 &   5170.35  &   7.921982  & -24.9336   &   5237.29   &  5.693702  &  -24.9434 \\
13 &   5165.73  &   7.923401  &  -1.5973   &   5232      &  5.702844  &   -1.5973 \\
14 &   5110.57  &   7.924798  &   2.8806   &   5186.13   &  5.712348  &    2.8805 \\
15 &   5050.42  &   7.926171  &  -1.6368   &   5050.87   &  5.721648  &   -1.6368 \\
16 &   4878.18  &   7.927546  &   7.2445   &   4884.18   &  5.728177  &    7.2466 \\
17 &   4796.39  &   7.928953  & -12.8362   &   4761.34   &  5.737326  &  -12.8376 \\
18 &    4735.7  &   7.930365  & -25.2287   &   4705.51   &  5.743843  &  -25.2428 \\
19 &   4688.86  &   7.931775  &  -1.3620   &   4662.16   &  5.751742  &   -1.3618 \\
20 &   4609.26  &   7.933138  &   2.8029   &   4530.53   &  5.760916  &    2.8027 \\
21 &   4505.35  &   7.934551  &  -1.3920   &   4465.29   &  5.767511  &   -1.3918 \\
22 &   4466.61  &   7.935955  &  10.6028   &   4379.13   &  5.775407  &   10.6040 \\
23 &   4373.21  &   7.937358  &  -9.5433   &   4323.05   &  5.783387  &   -9.5435 \\
24 &   4266.42  &   7.938775  & -17.4184   &   4191.96   &  5.792649  &  -17.4255 \\
25 &   4220.26  &   7.940171  &  -1.3810   &   4136.08   &  5.803394  &   -1.3805 \\
26 &   4050.48  &   7.941566  &   2.8130   &   4044.95   &  5.808722  &    2.8127 \\
27 &   4023.58  &   7.942962  &  -1.4007   &   3938.68   &  5.815519  &   -1.4002 \\
28 &   3992.68  &   7.944357  &  10.2221   &   3916.47   &  5.824949  &   10.2234 \\
29 &    3767.4  &   7.945721  & -10.2300   &   3846.8    &  5.831830  &  -10.2301 \\
30 &   3709.22  &   7.947147  & -19.7626   &   3576      &  5.840273  &  -19.7701 \\
31 &   3614.32  &   7.948520  &  -0.8885   &   3536.47   &  5.849072  &   -0.8886 \\
32 &   3405.79  &   7.949917  &   2.6443   &   3349.57   &  5.856030  &    2.6440 \\
33 &   3191.31  &   7.951232  &  -0.8455   &   3205.81   &  5.860392  &   -0.8456 \\
34 &   3092.14  &   7.952565  &   8.7725   &   3169.47   &  5.864614  &    8.7735 \\
35 &   2669.06  &   7.953911  &  -9.1266   &   2766.8    &  5.867528  &   -9.1268 \\
36 &   1935.55  &   7.955214  & -17.3202   &   1872.21   &  5.868979  &  -17.3295 \\
37 &   1856.89  &   7.956483  &            &   1534.06   &  5.871888  &           \\
38 &   1406.16  &   7.957785  &            &   1422.71   &  5.873300  &           \\
39 &   1215.79  &   7.959123  &            &   1247.39   &  5.874730  &           \\
40 &   1146.05  &   7.960446  &            &   1163.82   &  5.877753  &           \\
41 &   1062.19  &   7.961723  &            &   1105.05   &  5.879195  &           \\
42 &   990.388  &   7.963026  &            &   1035.38   &  5.880571  &           \\
43 &   944.654  &   7.964327  &            &   970.948   &  5.881957  &           \\
44 &   896.425  &   7.965627  &            &   916.323   &  5.883337  &           \\
45 &   868.715  &   7.966990  &            &   886.546   &  5.884838  &           \\
46 &   854.028  &   7.969591  &            &   864.105   &  5.886252  &           \\
47 &   846.043  &   7.970898  &            &   853.08    &  5.889118  &           \\
48 &   843.861  &   7.972203  &            &   848.483   &  5.890505  &           \\
49 &   842.991  &   7.973501  &            &   844.994   &  5.891939  &           \\
50 &     841.9  &   7.974808  &            &   843.241   &  5.893339  &           \\
51 &   841.566  &   7.976111  &            &   842.484   &  5.894753  &           \\
52 &   841.119  &   7.977419  &            &   841.523   &  5.896165  &           \\
53 &   840.647  &   7.978721  &            &   840.877   &  5.897563  &           \\
54 &   840.618  &   7.980032  &            &   840.686   &  5.898996  &           \\
55 &   840.558  &   7.981342  &            &   840.584   &  5.900508  &           \\
56 &   840.542  &   7.982662  &            &   840.515   &  5.901898  &           \\
57 &   840.542  &   7.984747  &            &   840.511   &  5.903295  &           \\
58 &{\bf 840.542}&  7.985218  &  $\leftarrow${\bf 0.0079997} &{\bf 840.511}&  5.914601   &$\leftarrow${\bf 0.0306858}	    \\\hline
\end {tabular}}
\end{table}}

\section{Raw Data of the Extracted Feature Locations}
Please see corners.dat.

\section{Matlab Source Code}
Please see codes.m.

\clearpage
\bibliography{D:/work/cameracalibration/calibration}

\begin{thebibliography}{10}

\bibitem{zhang99calibrationinpaper}
Zhengyou Zhang,
\newblock ``Flexible camera calibration by viewing a plane from unknown
  orientation,''
\newblock {\em IEEE International Conference on Computer Vision}, pp. 666--673,
  Sep. 1999.

\bibitem{zhang99calibrationinreport}
Zhengyou Zhang,
\newblock ``A flexible new technique for camera calibration,'' Microsoft
  Research Technical Report, {\tt http://}{\tt research.} {\tt microsoft}{\tt
  .com/}{\tt \char126}{\tt zhang/calib/}, 1998.

\bibitem{Richard97indefenseof8-pointalgorithm}
Richard~I. Hartley,
\newblock ``In defense of 8-point algorithm,''
\newblock {\em IEEE Trans. on Pattern Analysis and Machine Intelligence}, vol.
  19, no. 6, pp. 580--593, June 1997.

\bibitem{Heikkil97fourstepcameracalibration}
J.~Heikkil and O.~Silvn,
\newblock ``A four-step camera calibration procedure with implicit image
  correction,''
\newblock in {\em IEEE Computer Society Conference on Computer Vision and
  Pattern Recognition}, San Juan, Puerto Rico, 1997, pp. 1106--1112.

\bibitem{STURM99planebasedcalibrationsigularities}
P.~Sturm and S.~Maybank,
\newblock ``On plane-based camera calibration: a general algorithm,
  singularities, applications,''
\newblock {\em Proceedings of the Conference on Computer Vision and Pattern
  Recognition}, pp. 432--437, June 1999.

\bibitem{intel}
Online Document,
\newblock ``Camera calibration toolbox for {M}atlab,'' {\tt
  http://www.vision.caltech} {\tt .edu/bouguetj/calib\_doc/}.

\bibitem{Seth96visualservoing}
Seth Hutchinson, Gregory~D. Hager, and Peter~I. Corke,
\newblock ``A tutorial on visual servo control,''
\newblock {\em IEEE Transactions on Robotics and Automation}, vol. 12, no. 5,
  pp. 651--670, Sep. 1996.

\bibitem{Emanuele98introductorycomputervision}
Emanuele Trucco and Alessandro Verri,
\newblock {\em Introductory Techniques for 3-D Computer Vision},
\newblock Prentice Hall, 1998.

\bibitem{Richard94robotics}
Richard~M. Murray, Zexiang Li, and S.~Shankar Sastry,
\newblock {\em A Mathematical Introduction to Robotic Manipulation},
\newblock CRC Press, 1994.

\bibitem{Juyang92distortionmodel}
Juyang Weng, Paul Cohen, and Marc Herniou,
\newblock ``Camera calibration with distortion models and accuracy
  evaluation,''
\newblock {\em IEEE Transactions on Pattern Analysis and Machine Intelligence},
  vol. 14, no. 10, pp. 965--980, Oct. 1992.

\bibitem{Rudyreport}
Rudy,
\newblock ``{ODIS} and the use of computer vision,'' CSOIS Technical Report,
  Department of Electrical and Computer Engineering, Utah State University,
  2001.

\bibitem{Lili01connectedcomponentlabeling}
Lili Ma,
\newblock ``Localization using yellow line,'' CSOIS Technical Report,
  Department of Electrical and Computer Engineering, Utah State University,
  2001.

\bibitem{Gerhard00HandbookofComputerVisionAlgorithmsinImageAlgebra}
Gerhard~X. Ritter and Joseph~N. Wilson,
\newblock {\em Handbook of Computer Vision Algorithms in Image Algebra},
\newblock CRC Press, 2000.

\bibitem{Katsoulas01scanlineapproximation}
D.~K. Katsoulas and D.~I. Kosmopoulos,
\newblock ``An efficient depalletizing system based on 2{D} range imagery,''
\newblock in {\em Proceedings of the IEEE International Conference on Robotics
  and Automation}. IEEE, 2001, pp. 305--312.

\bibitem{Lili02Fit}
N.~S. Flann, K.~L. Moore, and L.~Ma,
\newblock ``A small mobile robot for security and inspection operations,''
\newblock in {\em Proceedings of 1st IFAC Conference on Telematics Applications
  in Automation and Robotics}, Weingarten, Germany, July 2001, IFAC, pp. 1--6.

\bibitem{zhang98calibrationwebpage}
Zhengyou Zhang,
\newblock ``Experimental data and result for camera calibration,'' Microsoft
  Research Technical Report, {\tt http://}{\tt rese-} {\tt arch.}{\tt
  microsoft.com/}{\tt \char126 zhang/calib/}, 1998.

\bibitem{odiscamera}
``Cm3000-l29 color board camera ({ODIS} camera) specification sheet,'' {\tt
  http}{\tt://www.}{\tt video-}{\tt surveillance}{\tt-hidden}{\tt-spy}
  {\tt-cameras}{\tt .com/}{\tt cm3000l29.htm}.

\bibitem{chapter4}
Paul~Ernest Debevec,
\newblock {\em Modeling and rendering architecture from photographs},
\newblock Ph.D. thesis, Computer Science Department, University of Michigan at
  Ann Arbor, 1996.

\bibitem{finalreportinStanford}
Michal Smulski and Mai Vu,
\newblock ``Autonomous calibration toolbox for cmos camera arrays,'' {\tt
  http://www.stanford.edu/\char126 mhv/cs233b/report.html}.

\bibitem{Song02sparsehough}
Zhen Song, YangQuan Chen, Lili Ma, and You~Chung Chung,
\newblock ``Some sensing and perception techniques for an omni-directional
  ground vehicles with a laser scanner,''
\newblock in {\em Proceedings of the 17th IEEE International Symposium on
  Intelligent Control, IEEE ISIC'02}, Vancouver, British Columbia, Canada, Oct.
  2002, IEEE, pp. 1--6.

\bibitem{Milan98imageprocessinganalysisandmachinevision}
Milan Sonka, Vaclav Hlavac, and Roger Boyle,
\newblock {\em Image Processing, Analysis, and Machine Vision},
\newblock PWS Publishing, 1998.

\bibitem{Undistortionchapter}
Charles Lee,
\newblock {\em Radial Undistortion and Calibration on An Image Array},
\newblock Ph.D. thesis, MIT, 2000.

\bibitem{berkemeier01visualservo}
Matthew Berkemeier, Morgan Davidson, Vikas Bahl, YangQuan Chen, and Lili Ma,
\newblock ``Visual servoing of an omnidirectional mobile robot for alignment
  with parking lot lines,''
\newblock in {\em Proceedings of the IEEE Int. Conf. Robotics and Automation
  (ICRA'02)}. IEEE, 2002, pp. 4204--4210.

\bibitem{lili02visualservo}
Lili Ma, Matthew Berkemeier, YangQuan Chen, Morgan Davidson, and Vikas Bahl,
\newblock ``Wireless visual servoing for {ODIS}: an under car inspection mobile
  robot,''
\newblock in {\em Proceedings of the 15th IFAC Congress}. IFA, 2002, pp.
  21--26.

\bibitem{TOMLABmanual1}
Kenneth Holmstrom and Mattias Bjorkman,
\newblock ``The {TOMLAB NLPLIB} toolbox for nonlinear programming,''
\newblock {\em Advanced Modeling and Optimization}, vol. 1, no. 1, 1999.

\bibitem{TOMLABmanual2}
Kenneth Holmstrom,
\newblock ``The {TOMLAB} optimization environment in {M}atlab,''
\newblock {\em Advanced Modeling and Optimization}, vol. 1, no. 1, 1999.

\bibitem{MATLABoptimizationtoolbox}
``Matlab optimization toolbox user's guide,'' {\tt http://www.mathworks.com}.

\end{thebibliography}
\end{document}